\pdfoutput=1

\documentclass[11pt]{article}

\usepackage[preprint]{coling}

\usepackage{times}
\usepackage{latexsym}

\usepackage[T1]{fontenc}

\usepackage[utf8]{inputenc}

\usepackage{microtype}

\usepackage{inconsolata}

\usepackage{graphicx}
\usepackage{url}
\usepackage{algorithm}
\usepackage{algorithmic}
\usepackage{enumitem}
\usepackage{tikz}
\definecolor{light_gray}{HTML}{FFFFFF} 
\definecolor{solid_gray}{HTML}{000000} 
\usepackage{multirow}
\usepackage{array}
\usepackage{hhline}
\usepackage{pifont}
\usepackage{makecell}
\usepackage{array} 
\usepackage{booktabs}
\usepackage{siunitx}
\usepackage{nameref}
\usepackage{xcolor}
\usepackage{transparent}
\usepackage{soul}
\usetikzlibrary{calc}
\usepackage{tcolorbox}
\usepackage{amssymb}
\usepackage{bbding}
\usepackage{pifont}
\usepackage{longtable}
\usepackage{tabularx}
\usepackage{rotating}
\usepackage{adjustbox}
\usepackage{bibentry}
\usepackage{pgfplots}
\usepackage{amsmath}
\usepackage{pgfplotstable}
\usepackage{comment}
\usepackage{float} 
\usetikzlibrary{positioning} 

\usepackage{tikz}
\usetikzlibrary{shapes}

\newtcbox{\highlight}[1][]{%
    colback=green!15!white,
    colframe=green!15!white,
    boxrule=0pt,
    boxsep=0pt,
    left=2pt,
    right=2pt,
    top=2pt,
    bottom=2pt,
    sharp corners,
    #1
}

\definecolor{transblue}{rgb}{0.69, 0.85, 0.96} 
\definecolor{editcolor}{rgb}{0.7, 0, 0} 

\newtcbox{\bleuhl}[1][]{%
    on line,
    arc=0pt,
    outer arc=0pt,
    colback=transblue!50,
    boxsep=0pt,
    left=2.0pt,
    right=2.0pt,
    top=1.0pt,
    bottom=0.4pt,
    boxrule=0.0pt,
    #1
}

\title{Grading Massive Open Online Courses Using Large Language Models}

\author{Shahriar Golchin$^\dagger$\thanks{Corresponding author.}, Nikhil Garuda$^\diamond$, Christopher Impey$^\diamond$, Matthew Wenger$^\diamond$ \\
  $^\dagger$Department of Computer Science, University of Arizona \\
  $^\diamond$Department of Astronomy, University of Arizona \\
  \texttt{golchin@arizona.edu} \\\\}

\begin{document}
\maketitle
\begin{abstract}
Massive open online courses (MOOCs) offer free education globally. Despite this democratization of learning, the massive enrollment in these courses makes it impractical for an instructor to assess every student’s writing assignment. As a result, peer grading, often guided by a straightforward rubric, is the method of choice. While convenient, peer grading often falls short in terms of reliability and validity. In this study, we explore the feasibility of using large language models (LLMs) to \textit{replace peer grading in MOOCs.}
To this end, we adapt the zero-shot chain-of-thought (ZCoT) prompting technique to \textit{automate the feedback process} once the LLM assigns a score to an assignment.
Specifically, to instruct LLMs for grading, we use three distinct prompts based on ZCoT: (1) ZCoT with instructor-provided correct answers, (2) ZCoT with both instructor-provided correct answers and rubrics, and (3) ZCoT with instructor-provided correct answers and LLM-generated rubrics. 
We tested these prompts in 18 different scenarios using two LLMs—GPT-4 and GPT-3.5—across three MOOCs: Introductory Astronomy, Astrobiology, and the History and Philosophy of Astronomy.
Our results show that ZCoT, when augmented with instructor-provided correct answers and rubrics, produces grades that are more aligned with those assigned by instructors compared to peer grading.
Finally, our findings indicate a promising potential for automated grading systems in MOOCs, especially in subjects with well-defined rubrics, to improve the learning experience for millions of online learners worldwide.
\end{abstract}


\section{Introduction}

Massive open online courses (MOOCs)\footnote{\url{https://www.mooc.org/}} have revolutionized access to free education by offering a wide array of free courses to anyone with an internet connection. This platform has democratized education and made it globally accessible, accommodating an unlimited number of participants \citep{impey2016bringing,Impey_Wenger_Austin_2015}.
However, the resulting increase in participant numbers, though beneficial to a broader audience, introduces challenges in delivering personalized and constructive feedback on student performance, including the task of grading their assignments. To address this issue, peer grading has become a common practice. While peer grading enhances students' engagement and motivation, its reliability and validity are often questioned \cite{FORMANEK2017243}.

In this study, we investigate the potential of LLMs in handling the task of grading assignments in MOOCs and the viability of replacing the current peer grading system with LLMs on this platform, aiming to establish more personalized and automated grading feedback to enhance the learning experience for online learners.
Specifically, our methodology is grounded on employing the zero-shot chain-of-thought (ZCoT) prompting technique by asking the LLMs to grade assignments ``step by step'' \cite{NEURIPS2022_8bb0d291}. 
Although this technique was originally designed to enhance the reasoning abilities of LLMs, we adapt it to generate feedback to students by having the LLMs verbalize the reasons behind score deductions.
We also further refine it by providing more in-context information about the given task.
In particular, we augment this technique in three ways: (1) ZCoT with instructor-provided correct answers, (2) ZCoT integrated with both instructor-provided correct answers and rubrics, and (3) ZCoT coupled with instructor-provided correct answers and LLM-generated rubrics based on the correct answers.
We use these prompting strategies to evaluate the grading performance of two performant LLMs—GPT-4 \citep{DBLP:journals/corr/abs-2303-08774} and GPT-3.5 \citep{DBLP:conf/nips/Ouyang0JAWMZASR22}—in three MOOC subjects: Astrobiology \citep{impey_astrobiology}, Introductory Astronomy \citep{impey_astronomy}, and the History and Philosophy of Astronomy \citep{impey_history}.
Our findings suggest that GPT-4 using ZCoT with instructor-provided answers and rubrics outperforms peer grading in terms of score alignment with those given by instructors.
However, grading courses that require imaginative or speculative thinking, such as History and Philosophy of Astronomy, is found to be a challenge for both LLMs and peer grading. Despite this, GPT-4 still excels over peer grading by producing scores more closely aligned with those assigned by the instructors in such scenarios.

The key contributions of this paper are:

\noindent \textbf{(1)} We study the potential of replacing the peer grading system with the use of LLMs and adapting the ZCoT prompting technique in MOOCs, aiming to provide automated grading feedback that is more accurate and reliable than peer grading.

\noindent \textbf{(2)} Using a rigorous evaluation across 18 settings, our results reveal that when integrating ZCoT with instructor-provided correct answers and rubrics, along with GPT-4, it outperforms all other settings, including peer grading. This approach results in grades that are more closely aligned with those provided by instructors and offers more useful feedback than peer grading. However, the score alignment is stronger in courses that require less creative and imaginative thinking.

\section{Related Work}

\textbf{LLMs in Education.}
The use of LLMs in education has primarily focused on how instructors and students leverage them to generate and respond to educational content \citep{Kasneci_2023,Alseddiqi_2023}. A key example is the implementation of GPT-4 at Khan Academy, where it acts as a learning assistant, answering subject-specific questions \citep{khan_openai}.
Recent studies expanded on this, exploring LLMs' capabilities in providing feedback on open-ended questions and essays \citep{matelsky2023large,Pinto_2023,meyer2024using,stahl2024exploring}, grading short-answer questions \citep{chang2024automatic}, generating programming exercises \citep{10.1145/3626252.3630863}, and helping students with coding questions \citep{10578705} or offering feedback on their code \citep{pankiewicz2023large,10343457,zhang2024students,gabbay2024combining,heickal2024generating,jacobs2024evaluating}.
Before LLMs became widely used, research in this area relied on the encoder-only component of Transformer networks \citep{devlin2018bert,vaswani2017attention} to evaluate assignments in short question-answering contexts \citep{Morris_2023,Hiroaki_2022,Riordan_2017}. 

\vspace{0.1cm}

\noindent\textbf{LLMs as Judges.} The most common use of LLMs is to approximate human judgment. Beyond grading, LLMs are used in various tasks requiring evaluative decisions. Among many examples, they were used to assess the quality of generated outputs in generation tasks such as summarization \citep{liu2023g}, evaluate translation quality \citep{kocmi-federmann-2023-large}, and detect factual inconsistencies \citep{luo2023chatgpt}.
LLMs were also applied to assess how well other LLMs follow instructions \citep{zeng2023evaluating}, gauge the similarity between reference and LLM-generated texts \citep{DBLP:journals/corr/abs-2308-08493}, and evaluate code \citep{zhuo2023ice}. Building on this, fine-tuned LLMs as judges were proposed to enhance the performance of general-purpose LLMs \citep{zhu2023judgelm}.

\vspace{0.08cm}

\noindent This research builds on earlier studies of peer grading in MOOCs, which revealed challenges in achieving consistency, reliability, and validity in peer-assigned grades \citep{FORMANEK2017243,Gamage_2021}.
Our study explores the potential of using LLMs to replace peer grading in MOOCs, aiming to improve and automate the feedback and grading process.

\section{Approach}
\subsection{Prompts}
To guide LLMs in grading assignments, we use zero-shot chain-of-thought (ZCoT) \citep{NEURIPS2022_8bb0d291} as our base prompting technique.
There are two main reasons behind this choice.
First, we performed a pilot study and found that ZCoT yields scores more closely aligned with those given by instructors compared to vanilla zero-shot prompting.
Second, ZCoT enables us to observe the reasoning behind each generated score, as the LLM is required to verbalize its reasoning process ``step by step.'' This allows us to analyze the reliability and validity of the grading steps provided as feedback to students and proactively mitigate potential hallucinations during the grading process.
Note that our use of ZCoT differs from its conventional application. Typically, ZCoT involves reasoning before making predictions. However, in our approach, the score (prediction) is generated first, followed by reasoning as feedback on any score deductions. Thus, we use ZCoT not to enhance performance, but to automatically generate feedback.




We develop three different prompts, which incorporate ZCoT in combination with: (1) instructor-provided correct answers, (2) instructor-provided correct answers and rubrics, and (3) instructor-provided correct answers with LLM-generated rubrics.
Each prompt is detailed in the following.

\begin{figure*}[!t]
    \begin{minipage}{\textwidth} 
        \centering
        \begin{tikzpicture}[rounded corners=8pt, thick, text=black, text opacity=1]
            \node[draw=solid_gray, fill=light_gray, line width=1pt, 
            text=black, text width=0.97\textwidth, align=left, 
            font=\fontsize{7pt}{9.5pt}\selectfont, inner xsep=5pt, 
            inner ysep=5pt] at (0,0) {\textbf{Instruction:}
            You are a fair and knowledgeable instructor whose task is to evaluate the 
            student's assignments in accordance with the correct answers to each of the 
            questions that are presented in the section that follows.
            Make sure you speak your thoughts aloud so that the students can understand 
            the rationale for any points deducted.
            Let's grade assignments step by step.

            -- -- --

            \textbf{Questions/Answers:}

            \vspace{0.05cm}

            \textbf{Questions:}

            What are the advantages of large telescopes? Provide at least one.
            
            Why do astronomers want telescopes in space when putting them there is expensive?
            
            What are some examples of wavelength regions beyond the spectrum of visible light 
            where astronomers can learn about the universe? Provide at least two.

            \vspace{0.05cm}

            \textbf{Answers:}

            Large telescopes collect more light and so permit fainter and more distant objects 
            to be seen. Also, large telescopes also generally have higher angular resolution, 
            although realizing this depends on being able to correct for blurring by the Earth's 
            atmosphere. 
            Telescopes in space are immune from the blurring effect of the Earth's atmosphere, 
            which reduces the depth of observation and limits the ability to resolve or separate 
            object close on the sky.
            For some wavelength regimes beyond the visible spectrum, 
            astronomy must be done from space because those wavelengths do not penetrate the 
            Earth's atmosphere. Telescopes in space are also immune from light pollution on 
            Earth.
            Wavelength ranges include: radio, microwave, infrared, ultraviolet, X-ray, and gamma 
            ray.

            \vspace{0.05cm}

            \textbf{Grade: 9/9}

            -- -- --

            \textbf{Student's Answers:}

           Astronomers can learn about the universe by investigating certain wavelengths such as cosmic and gamma rays, X-rays, infrared, and radio waves. For example, the Cosmic Background Radiation (CMB) can tell astronomers something about the recombination epoch. It is very important for observational cosmology.

            X-rays are produced in the universe when matter is heated to millions of degrees. These are therefore used to detect areas with high magnetic fields and extreme gravity plus areas where huge cosmic explosions occur.
            
            Likewise, gamma ray bursts (GRBs) occur when a massive star collapses to form a neutron star or black hole.
            [...]
                        
            \textit{\{The answer is partially cut due to space constraints.\}}

            \textbf{Grade: }

            -- -- --
            
            \vspace{-0.1cm}
            
            \highlight{
                7/9
                }
            \vspace{-0.315cm}
            \highlight{The student’s answer is quite comprehensive and shows a good understanding of the subject matter. However, the student did not directly answer}
            \vspace{-0.3cm}
            \highlight{the first question about the advantages of large telescopes. While they did mention that “The light-gathering power and the resolution of a telescope}
            \vspace{-0.3cm}
            \highlight{increase with increasing aperture”, they did not explicitly state that these are advantages of large telescopes. The student did a good job explaining why}
            \vspace{-0.3cm}
            \highlight{astronomers want telescopes in space, mentioning that they ``give ultra-sharp optical images and to detect long and short wavelengths that cannot}
            \vspace{-0.3cm}
            \highlight{penetrate the Earth’s atmosphere.” The student also did not provide specific examples of wavelength regions beyond the spectrum of visible light where}
            \vspace{-0.3cm}
            \highlight{astronomers can learn about the universe, which was asked in the third question. They mentioned the “peeling back of the electromagnetic spectrum,}
            \vspace{-0.3cm}
            \highlight{revealing for the first time the invisible universe” but did not provide specific examples as requested. Therefore, 2 points are deducted for not directly}
            \vspace{-0.3cm}
            \highlight{answering the first question and not providing specific examples for the third question.}

            };
        \end{tikzpicture}
    \end{minipage}
    \caption{An illustration of the ZCoT prompt along with answers provided by the course instructor. Each question is assessed individually for every student. We repeat this process for all questions and students, incorporating their answers into the prompt, and instructing the LLM to grade the assignments. In this example, the instructor-assigned grade is 6/9, with GPT-4 serving as the underlying LLM.}
    \label{figure:ZCoT-with-answers}
\end{figure*}

\vspace{0.08cm}

\noindent\textbf{ZCoT with Instructor-Provided Correct Answers.} Initially, we implement ZCoT by incorporating the correct answers, supplied by the course instructor, as in-context information to generate grades for assignments. Further, each student's assignment is embedded into the input prompt, and the LLM is tasked with grading the assignment based on the instructor-provided answer for each question.
Note that, in this prompt, there is no guidance on how the grading should be carried out, except for the total points for each question. Finally, the grading is conducted for each student and question separately, ensuring that there is no influence from other questions or the grades of other students on the downstream grading task. Figure \ref{figure:ZCoT-with-answers} shows the ZCoT with correct answers integrated.

\begin{figure*}[!t]
    \begin{minipage}{\textwidth} 
        \centering
        \begin{tikzpicture}[rounded corners=8pt, thick, text=black, text opacity=1]
            \node[draw=solid_gray, fill=light_gray, line width=1pt, 
            text=black, text width=0.97\textwidth, align=left, 
            font=\fontsize{7pt}{9.5pt}\selectfont, inner xsep=5pt, 
            inner ysep=5pt] at (0,0) {\textbf{Instruction:}
            You are a fair and knowledgeable instructor whose task is to 
            evaluate the student's assignments in accordance with the correct 
            answers and provided rubric to each of the questions that are 
            presented in the section that follows.
            
            Make sure that the score deduction follows the provided rubric.
            Let's grade assignments step by step and explain the reasons behind the point 
            deduction given the rubric.

            -- -- --

            \textbf{Rubrics:}



            Does the writer answer the question correctly and express their thoughts clearly using supporting facts and relevant examples?
            
            3 points: The writer answers the question correctly and the writer expresses their thoughts clearly, and uses supporting facts and relevant examples
            
            2 points: The writer answers the question correctly, but does not explain it well.
            
            0 points: The writer does not answer the question correctly

            \vspace{0.1cm}

            Does the writer answer the question correctly and express their thoughts clearly using supporting facts and relevant examples?
            
            3 points: The writer answers the question correctly and the writer expresses their thoughts clearly, and uses supporting facts and relevant examples
            
            2 points: The writer answers the question correctly, but does not explain it well.
            
            0 points: The writer does not answer the question correctly

            \vspace{0.1cm}

            Does the writer answer the question correctly and the writer expresses their thoughts clearly using supporting facts and relevant examples? 
            
            3 points: The writer answers the question correctly and the writer expresses their thoughts clearly, and uses supporting facts and relevant examples
            
            2 points: The writer answers the question correctly, but does not explain it well. The write only includes one wavelength instead of two.
            
            0 points: The writer does not answer the question correctly
            
            \vspace{0.05cm}

            \textbf{Questions/Answers:}

            \textit{\{As Figure \ref{figure:ZCoT-with-answers}.\}}

            \vspace{0.05cm}

            \textbf{Grade: 9/9}

            -- -- --

            \textbf{Student's Answers:}

            \textit{\{As Figure \ref{figure:ZCoT-with-answers}.\}}


            \vspace{0.05cm}

            \textbf{Grade:}

            -- -- --

            \vspace{-0.10cm}
            
            \highlight{6/9}
            \vspace{-0.3cm}
            \highlight{The student does not directly answer the first question about the advantages of large telescopes. While they mention that telescopes improve on the}
            \vspace{-0.3cm}
            \highlight{light-gathering power of the eye and allow astronomers to resolve finer details, they do not specifically state these as advantages of large telescopes.}
            \vspace{-0.3cm}
            \highlight{This is worth 2 points instead of 3.}
            \vspace{-0.3cm}
            \highlight{The student answers the second question correctly and explains well why astronomers want telescopes in space. They mention that telescopes in space}
            \vspace{-0.3cm}
            \highlight{can give ultra-sharp optical images and detect long and short wavelengths that cannot penetrate the Earth’s atmosphere. This is worth 3 points.}
            \vspace{-0.3cm}
            \highlight{The student does not directly answer the third question about examples of wavelength regions beyond the spectrum of visible light. While they} 
            \vspace{-0.3cm}
            \highlight{mention the peeling back of the electromagnetic spectrum, they do not provide specific examples of wavelength regions. This is worth 1 point instead} 
            \vspace{-0.3cm}
            \highlight{of 3.} 
            };
        \end{tikzpicture}
    \end{minipage}
    \caption{An illustration of ZCoT prompt that incorporates both instructor-provided correct answers and rubrics for grading assignments. The grading process used in ZCoT with correct answers only (Figure \ref{figure:ZCoT-with-answers}) is also applied here. Similar to Figure \ref{figure:ZCoT-with-answers}, the instructor-assigned grade for this question is 6/9, and GPT-4 is the base model. As shown, including rubrics in the prompt helps the LLM generate a grade consistent with the grade assigned by the instructor.}
    \label{figure:ZCoT-with-answers-and-rubrics}
\end{figure*}

\begin{figure*}[!t]
    \begin{minipage}{\textwidth} 
        \centering
        \begin{tikzpicture}[rounded corners=8pt, thick, text=black, text opacity=1]
            \node[draw=solid_gray, fill=light_gray, line width=1pt, 
            text=black, text width=0.97\textwidth, align=left, 
            font=\fontsize{7pt}{9.5pt}\selectfont, inner xsep=5pt, 
            inner ysep=5pt] at (0,0) {\textbf{Instruction:}
            Your task is to design a rubric that addresses the following questions/answers. 
            The followings are the homework assignments for the Astrobiology course. 
            This course is designed for undergraduate students majoring in Astronomy. 
            The rubric must be unbiased and adaptable, capable of fairly evaluating any 
            kind of student writing assignment.

            The scoring breakdown for each question should be as follows: the score for 
            Question 1 is 10, the score for Question 2 is 10, and the score for Question 3 
            is 10. Your thoughtful consideration in creating this rubric will 
            ensure all students' work is evaluated equitably and consistently. Make sure
            that the rubric will provide points that are whole numbered.

            Each rubric should be dedicated to each question separately.

            -- -- --
            
            \textbf{Questions/Answers:}

            \vspace{0.05cm}

            \textbf{Question 1:}

            \textit{\{As Figure \ref{figure:ZCoT-with-answers}.\}}

            \vspace{0.05cm}

            \textbf{Answer 1:}

            \textit{\{As Figure \ref{figure:ZCoT-with-answers}.\}}

            Full Score: 10/10

            \vspace{0.05cm}
            
            \textbf{Question 2:}

            Discuss how habitable zone range and spectral type are related. [...]

            \textit{\{The question is partially cut due to space constraints.\}}

            \vspace{0.05cm}

            \textbf{Answer 2:}

            The habitable zone is defined as the range of distances from a star within which liquid water can exist on a planet. [...]

            \textit{\{The answer is partially cut due to space constraints.\}}

            Full Score: 10/10

            \vspace{0.05cm}

            \textbf{Question 3:}

            Discuss why/whether it is plausible that exobiology exists at all, 
            in terms of exoplanet type, spectral type and orbital distance. [...]

            \textit{\{The question is partially cut due to space constraints.\}}

            \vspace{0.05cm}

            \textbf{Answer 3:}

            The star around HD-127688 has a K spectral type.  [...]
            
           \textit{\{The answer is partially cut due to space constraints.\}}
            

            Full Score: 10/10

            -- -- --
            
            \vspace{-0.10cm}

            \highlight{Rubric for Question 1:}
            \vspace{-0.3cm}
            \highlight{- Identification of detection methods (2 points): The student correctly identifies the detection methods used for each exoplanet.}
            \vspace{-0.3cm}
            \highlight{- Explanation of detection methods (2 points): The student provides a clear and accurate explanation of how each detection method works.}
            \vspace{-0.3cm}
            \highlight{- Identification of physical characteristics (2 points): The student correctly identifies the physical characteristics that can be learned from each set of}
            \vspace{-0.3cm}
            \highlight{data.}
            \vspace{-0.3cm}
            \highlight{- Explanation of physical characteristics (2 points): The student provides a clear and accurate explanation of why these physical characteristics can be}
            \vspace{-0.3cm}
            \highlight{learned from the data.}
            \vspace{-0.3cm}
            \highlight{- Identification of Earth-like exoplanet (2 points): The student correctly identifies one exoplanet as Earth-like. [...]}
            
            
            \textit{\{The answer is partially cut due to space constraints.\}}

            };
        \end{tikzpicture}
    \end{minipage}
    \caption{An illustration of the prompt utilized to generate rubrics using GPT-4 for the Astrobiology course. This procedure is repeated for all courses under study, substituting the course name, correct answers, total grades, and questions accordingly. The generated rubrics are then integrated into the ZCoT prompt along with the correct answers for assignment grading. Specifically, the prompt showcased in Figure \ref{figure:ZCoT-with-answers-and-rubrics} is employed for grading, where the instructor-provided rubrics are replaced with rubrics generated by LLM.}
    \label{figure:template-prompt-for-LLM-generated-rubrics}
\end{figure*}

\vspace{0.08cm}

\noindent\textbf{ZCoT with Instructor-Provided Correct Answers and Rubrics.}
Our second prompting strategy enhances the previous approach by integrating the instructor-provided rubric for each question in addition to the correct answer. This method aims to align the LLM more closely with the instructor's grading criteria, ensuring the grades and score deductions match the instructor's standards.
Retaining all components from the prior prompting technique, we follow the same grading procedure as in the previous stage.
Figure \ref{figure:ZCoT-with-answers-and-rubrics} illustrates ZCoT with the inclusion of the correct answers and rubrics.

\vspace{0.08cm}

\noindent\textbf{ZCoT with Instructor-Provided Correct Answers and LLM-Generated Rubrics.}
In this approach, we diverge from the two previous methods where the reference information in the prompt is centered around the human (instructor), and instead, include a portion of information generated by the LLM. This idea is inspired by the fact that LLMs can generate better rubrics compared to those provided by an instructor. The reasoning behind this is that LLMs are trained on an extensive amount of data, and thus, possess comprehensive interdisciplinary knowledge, such as knowledge from both education and astronomy domains in our case, that allows them to produce improved rubrics. To implement this, we prompt GPT-4 with the template prompt shown in Figure \ref{figure:template-prompt-for-LLM-generated-rubrics} to generate rubrics based on the provided correct answer, total grade, and the question itself. This LLM-generated rubric is then replaced with the instructor-provided rubric used in the previous stage (Figure \ref{figure:ZCoT-with-answers-and-rubrics}), and the same process is repeated to grade all assignments.

\subsection{Evaluation of LLM-Assigned Grades}
\label{sec:eval-of-llm-assigned-grades}
As noted, MOOCs cater to a large number of participants for each course, and our study aims to evaluate the effectiveness of LLMs in grading assignments for a substantial student population. However, to control the budget when using proprietary LLMs, and carefully analyze the grades assigned by LLMs to individual questions, we conduct the grading process on a manageable subset of the entire collection of students' assignments (see Section \ref{section:experimental-setup} under Data for more details). On the other hand, it is crucial to analyze the performance of LLMs in grading a significant volume of assignments, representative of the regular situations in MOOCs. To address this, we use the bootstrap resampling \cite{10.1214/aos/1176344552,EfroTibs93}.

Qualitatively, we compare LLM-assigned grades with those given by instructors (ground truth). Specifically, we examine the \emph{absolute difference} in average grades per question for each course, considering both lenient and strict grading. The hypothesis we investigate here is that the difference between LLM-assigned and instructor-assigned grades should  \emph{not be statistically significant}, as determined through bootstrap resampling. In fact, this implies that the average grades given by both LLM and the instructor could be similar enough to be considered interchangeable.
In a different analysis, we also use bootstrap resampling to evaluate average scores for potential bias in LLM grading. This helps us understand how the \emph{overall average scores} from LLMs compare to those from instructors, providing insight into the reliability and consistency of LLM grading. In addition, we compute the mean absolute error (MAE) to measure the discrepancies between LLM- and instructor-assigned grades.

Quantitatively, we assess each question by comparing the LLM-assigned grades with the instructor-assigned grades using evaluations conducted by course instructors. A summary of this evaluation is then reported as our qualitative analysis.

\subsection{Baseline}
We analyze average scores and MAEs produced by the LLMs for each question across all courses, comparing them to the peer grading as the baseline and the instructor-assigned grades as the ground truth. By evaluating the average scores and MAEs with respect to the ground truth, our goal is to identify the most effective type of prompt and the corresponding LLM. In this context, a prompt, along with the respective LLM, is deemed superior if it demonstrates greater alignment with the instructor-assigned grades compared to those obtained through peer grading.

\begin{table*}[th]
    \centering
    \caption{Average grades from grading 10 students' writing assignments using the three proposed prompting techniques: ZCoT with instructor-provided correct answers, ZCoT combined with instructor-provided correct answers and rubrics, and ZCoT with instructor-provided correct answers and LLM-generated rubrics. Results encompass three MOOCs: Introductory Astronomy, Astrobiology, and the History and Philosophy of Astronomy. Grading is conducted using GPT-3.5 and GPT-4 as the underlying LLMs. Each average grade includes a subscripted standard deviation. As the differences in average grades for both peer and LLM grades relative to instructor grades are \emph{not statistically significant}, no grades are marked with an asterisk (see Appendix \ref{appendix:p_values} for details on $p$-values).}
    \label{tab:raw-resullts}
    \footnotesize

    \begin{adjustbox}{width=\textwidth,center}

        \begin{tabular}{@{}cccccccccc@{}}
            \toprule
                                                        &                                 &                                                                                        & \multicolumn{1}{l}{}                                                                & \multicolumn{3}{c}{\textbf{GPT-3.5}}                                                                                                                                                                                                                  & \multicolumn{3}{c}{\textbf{GPT-4}}                                                                                                                                                                                               \\ \cmidrule(lr){5-7}  \cmidrule(l){8-10}
            \multicolumn{1}{c|}{\textbf{Courses}}         & \multicolumn{1}{c|}{\textbf{\raisebox{-2.5ex}{\rotatebox{90}{Questions}}}} & \multicolumn{1}{c|}{\textbf{\begin{tabular}[c]{@{}c@{}}Instructor\\ Grades\end{tabular}}} & \multicolumn{1}{c}{\textbf{\begin{tabular}[c]{@{}c@{}}Peer\\ Grades\end{tabular}}} \vrule width 1pt & \multicolumn{1}{c|}{\textbf{\begin{tabular}[c]{@{}c@{}}ZCoT W/\\ Answers\end{tabular}}} & \multicolumn{1}{c|}{\textbf{\begin{tabular}[c]{@{}c@{}}ZCoT W/\\ Answers \&\\ Rubrics\end{tabular}}} & \multicolumn{1}{c}{\textbf{\begin{tabular}[c]{@{}c@{}}ZCoT W/\\ Answers \&\\ LLM Rubrics\end{tabular}}} \vrule width 1pt & \multicolumn{1}{c|}{\textbf{\begin{tabular}[c]{@{}c@{}}ZCoT W/\\ Answers\end{tabular}}} & \multicolumn{1}{c|}{\textbf{\begin{tabular}[c]{@{}c@{}}ZCoT W/\\ Answers \&\\ Rubrics\end{tabular}}} & \multicolumn{1}{c}{\textbf{\begin{tabular}[c]{@{}c@{}}ZCoT W/\\ Answers \&\\ LLM Rubrics\end{tabular}}} \\ \midrule
            \multicolumn{1}{c|}{\multirow{7}{*}{{\begin{tabular}[c]{@{}c@{}}Introductory \\ Astronomy\end{tabular}}}} & \multicolumn{1}{c|}{Q1}         & \multicolumn{1}{c|}{$3.90_{\pm 1.70}$}                                                              & \multicolumn{1}{c}{$5.15_{\pm 0.87}$} \vrule width 1pt                                                           & \multicolumn{1}{c|}{$4.50_{\pm 0.50}$}                                        & \multicolumn{1}{c|}{$2.90_{\pm 1.92}$}                                                        & \multicolumn{1}{c}{$4.10_{\pm 1.92}$} \vrule width 1pt                                                                          & \multicolumn{1}{c|}{$4.75_{\pm 1.29}$}                                       & \multicolumn{1}{c|}{$4.40_{\pm 1.28}$}                                                        & \multicolumn{1}{c}{$4.40_{\pm 1.50}$}                                                     \\ \cmidrule(l){2-10} 
            \multicolumn{1}{c|}{}                         & \multicolumn{1}{c|}{Q2}         & \multicolumn{1}{c|}{$8.20_{\pm 1.17}$}                                                               & \multicolumn{1}{c}{$7.55_{\pm 2.36}$} \vrule width 1pt                                                           & \multicolumn{1}{c|}{$7.60_{\pm 0.92}$}                                        & \multicolumn{1}{c|}{$8.30_{\pm 0.78}$}                                                        & \multicolumn{1}{c}{$8.40_{\pm 0.66}$} \vrule width 1pt                                                                          & \multicolumn{1}{c|}{$8.65_{\pm 0.63}$}                                        & \multicolumn{1}{c|}{$8.30_{\pm 0.90}$}                                                        & \multicolumn{1}{c}{$8.50_{\pm 0.87}$}                                                     \\ \cmidrule(l){2-10} 
            \multicolumn{1}{c|}{}                         & \multicolumn{1}{c|}{Q3}         & \multicolumn{1}{c|}{$7.50_{\pm 2.91}$}                                                               & \multicolumn{1}{c}{$7.40_{\pm 2.59}$} \vrule width 1pt                                                           & \multicolumn{1}{c|}{$6.80_{\pm 2.36}$}                                        & \multicolumn{1}{c|}{$7.20_{\pm 2.68}$}                                                        & \multicolumn{1}{c}{$6.60_{\pm 2.46}$} \vrule width 1pt                                                                          & \multicolumn{1}{c|}{$7.60_{\pm 2.73}$}                                        & \multicolumn{1}{c|}{$7.30_{\pm 2.72}$}                                                        & \multicolumn{1}{c}{$7.60_{\pm 2.75}$}                                                     \\ \cmidrule(l){2-10} 
            \multicolumn{1}{c|}{}                         & \multicolumn{1}{c|}{Q4}         & \multicolumn{1}{c|}{$7.40_{\pm 2.73}$}                                                               & \multicolumn{1}{c}{$7.45_{\pm 1.44}$} \vrule width 1pt                                                           & \multicolumn{1}{c|}{$6.80_{\pm 2.48}$}                                        & \multicolumn{1}{c|}{$7.10_{\pm 2.88}$}                                                        & \multicolumn{1}{c}{$6.80_{\pm 2.96}$} \vrule width 1pt                                                                          & \multicolumn{1}{c|}{$7.50_{\pm 2.42}$}                                        & \multicolumn{1}{c|}{$6.90_{\pm 2.84}$}                                                        & \multicolumn{1}{c}{$7.05_{\pm 2.81}$}                                                    \\ \cmidrule(l){2-10} 
            \multicolumn{1}{c|}{}                         & \multicolumn{1}{c|}{Q5}         & \multicolumn{1}{c|}{$5.50_{\pm 2.97}$}                                                               & \multicolumn{1}{c}{$7.40_{\pm 2.62}$} \vrule width 1pt                                                           & \multicolumn{1}{c|}{$6.40_{\pm 2.46}$}                                        & \multicolumn{1}{c|}{$7.40_{\pm 2.91}$}                                                        & \multicolumn{1}{c}{$6.40_{\pm 3.47}$} \vrule width 1pt                                                                          & \multicolumn{1}{c|}{$6.20_{\pm 2.86}$}                                        & \multicolumn{1}{c|}{$5.90_{\pm 3.36}$}                                                        & \multicolumn{1}{c}{$6.35_{\pm 3.49}$}                                                    \\ \midrule[0.9pt]
            \multicolumn{1}{c|}{\multirow{4}{*}{{\begin{tabular}[c]{@{}c@{}}Astrobiology\end{tabular}}}}   & \multicolumn{1}{c|}{Q1}         & \multicolumn{1}{c|}{$6.80_{\pm 3.49}$}                                                               & \multicolumn{1}{c}{$7.50_{\pm 2.54}$} \vrule width 1pt                                                            & \multicolumn{1}{c|}{$7.40_{\pm 2.65}$}                                        & \multicolumn{1}{c|}{$6.70_{\pm 3.74}$}                                                        & \multicolumn{1}{c}{$6.58_{\pm 2.98}$} \vrule width 1pt                                                                        & \multicolumn{1}{c|}{$7.50_{\pm 2.20}$}                                        & \multicolumn{1}{c|}{$7.10_{\pm 3.05}$}                                                        & \multicolumn{1}{c}{$7.10_{\pm 2.70}$}                                                     \\ \cmidrule(l){2-10} 
            \multicolumn{1}{c|}{}                         & \multicolumn{1}{c|}{Q2}         & \multicolumn{1}{c|}{$6.70_{\pm 2.65}$}                                                               & \multicolumn{1}{c}{$7.45_{\pm 3.24}$} \vrule width 1pt                                                           & \multicolumn{1}{c|}{$7.00_{\pm 2.41}$}                                        & \multicolumn{1}{c|}{$7.30_{\pm 2.65}$}                                                        & \multicolumn{1}{c}{$5.52_{\pm 4.11}$} \vrule width 1pt                                                                        & \multicolumn{1}{c|}{$7.90_{\pm 2.17}$}                                         & \multicolumn{1}{c|}{$7.40_{\pm 2.65}$}                                                        & \multicolumn{1}{c}{$7.10_{\pm 3.11}$}                                                     \\ \cmidrule(l){2-10} 
            \multicolumn{1}{c|}{}                         & \multicolumn{1}{c|}{Q3}         & \multicolumn{1}{c|}{$7.90_{\pm 2.70}$}                                                               & \multicolumn{1}{c}{$9.05_{\pm 1.65}$} \vrule width 1pt                                                           & \multicolumn{1}{c|}{$6.70_{\pm 2.28}$}                                        & \multicolumn{1}{c|}{$6.50_{\pm 2.69}$}                                                        & \multicolumn{1}{c}{$5.03_{\pm 4.22}$} \vrule width 1pt                                                                        & \multicolumn{1}{c|}{$8.10_{\pm 1.04}$}                                         & \multicolumn{1}{c|}{$7.50_{\pm 1.28}$}                                                        & \multicolumn{1}{c}{$7.50_{\pm 0.92}$}                                                     \\ \midrule[0.9pt]
\multicolumn{1}{c|}{\multirow{5.5}{*}{{\begin{tabular}[c]{@{}c@{}}History \& \\ Philosophy of \\ Astronomy \end{tabular}}}}     & \multicolumn{1}{c|}{Q1}                              & \multicolumn{1}{c|}{$3.50_{\pm 0.67}$}                                                                           & \multicolumn{1}{c}{$3.60_{\pm 0.66}$} \vrule width 1pt                                    & \multicolumn{1}{c|}{$2.70_{\pm 0.46}$}                                        & \multicolumn{1}{c|}{$2.00_{\pm 0.00}$}                                                          & \multicolumn{1}{c}{$2.85_{\pm 0.63}$} \vrule width 1pt                                                                       & \multicolumn{1}{c|}{$3.50_{\pm 0.67}$}                                                        & \multicolumn{1}{c|}{$3.20_{\pm 0.98}$}                                                        & \multicolumn{1}{c}{$3.20_{\pm 0.71}$}                                                     \\ \cmidrule(l){2-10} 
            \multicolumn{1}{c|}{}                         & \multicolumn{1}{c|}{Q2}         & \multicolumn{1}{c|}{$2.40_{\pm 0.92}$}                                                               & \multicolumn{1}{c}{$3.70_{\pm 0.64}$} \vrule width 1pt                                                           & \multicolumn{1}{c|}{$2.90_{\pm 0.70}$}                                        & \multicolumn{1}{c|}{$1.80_{\pm 0.98}$}                                                        & \multicolumn{1}{c}{$2.56_{\pm 0.78}$} \vrule width 1pt                                                                          & \multicolumn{1}{c|}{$3.25_{\pm 0.72}$}                                       & \multicolumn{1}{c|}{$3.10_{\pm 0.54}$}                                                        & \multicolumn{1}{c}{$2.95_{\pm 0.72}$}                                                    \\ \cmidrule(l){2-10} 
            \multicolumn{1}{c|}{}                         & \multicolumn{1}{c|}{Q3}         & \multicolumn{1}{c|}{$2.70_{\pm 0.64}$}                                                               & \multicolumn{1}{c}{$3.40_{\pm 1.20}$} \vrule width 1pt                                                           & \multicolumn{1}{c|}{$2.70_{\pm 1.00}$}                                        & \multicolumn{1}{c|}{$1.20_{\pm 1.33}$}                                                        & \multicolumn{1}{c}{$1.70_{\pm 1.19}$} \vrule width 1pt                                                                          & \multicolumn{1}{c|}{$3.65_{\pm 0.45}$}                                       & \multicolumn{1}{c|}{$3.20_{\pm 0.60}$}                                                        & \multicolumn{1}{c}{$3.20_{\pm 0.56}$}                                                     \\ \cmidrule(l){2-10} 
            \multicolumn{1}{c|}{}                         & \multicolumn{1}{c|}{Q4}         & \multicolumn{1}{c|}{$2.20_{\pm 0.87}$}                                                               & \multicolumn{1}{c}{$3.80_{\pm 0.60}$} \vrule width 1pt                                                           & \multicolumn{1}{c|}{$3.00_{\pm 0.45}$}                                        & \multicolumn{1}{c|}{$1.10_{\pm 1.04}$}                                                        & \multicolumn{1}{c}{$2.20_{\pm 0.60}$} \vrule width 1pt                                                                          & \multicolumn{1}{c|}{$3.25_{\pm 0.75}$}                                       & \multicolumn{1}{c|}{$2.70_{\pm 0.90}$}                                                        & \multicolumn{1}{c}{$2.95_{\pm 0.72}$}                                                                         \\ \bottomrule
        \end{tabular}
    \end{adjustbox}
\end{table*}

\begin{table*}[!th]
    \centering
        \caption{Average grades from grading 10 students' writing assignments using bootstrap resampled results in Table \ref{tab:raw-resullts} with 10,000 iterations. Each average grade includes a subscripted standard deviation.}
    \label{tab:bootstrap-results}
    \footnotesize
    \begin{adjustbox}{width=\textwidth,center}
        \begin{tabular}{@{}cccccccccc@{}}
            \toprule
                                                        &                                 &                                                                                        & \multicolumn{1}{l}{}                                                                & \multicolumn{3}{c}{\textbf{GPT-3.5}}                                                                                                                                                                                                                                                           & \multicolumn{3}{c}{\textbf{GPT-4}}                                                                                                                                                                                                                                        \\ \cmidrule(lr){5-7}  \cmidrule(l){8-10}
            \multicolumn{1}{c|}{\textbf{Courses}}         & \multicolumn{1}{c|}{\textbf{\raisebox{-2.5ex}{\rotatebox{90}{Questions}}}} & \multicolumn{1}{c|}{\textbf{\begin{tabular}[c]{@{}c@{}}Instructor\\ Grades\end{tabular}}} & \multicolumn{1}{c}{\textbf{\begin{tabular}[c]{@{}c@{}}Peer\\ Grades\end{tabular}}} \vrule width 1pt & \multicolumn{1}{c|}{\textbf{\begin{tabular}[c]{@{}c@{}}ZCoT W/\\ Answers\end{tabular}}} & \multicolumn{1}{c|}{\textbf{\begin{tabular}[c]{@{}c@{}}ZCoT W/\\ Answers \&\\ Rubrics\end{tabular}}} & \multicolumn{1}{c}{\textbf{\begin{tabular}[c]{@{}c@{}}ZCoT W/\\ Answers \&\\ LLM Rubrics\end{tabular}}} \vrule width 1pt & \multicolumn{1}{c|}{\textbf{\begin{tabular}[c]{@{}c@{}}ZCoT W/\\ Answers\end{tabular}}} & \multicolumn{1}{c|}{\textbf{\begin{tabular}[c]{@{}c@{}}ZCoT W/\\ Answers \&\\ Rubrics\end{tabular}}} & \textbf{\begin{tabular}[c]{@{}c@{}}ZCoT W/\\ Answers \&\\ LLM Rubrics\end{tabular}} \\ \midrule
            \multicolumn{1}{c|}{\multirow{7}{*}{{\begin{tabular}[c]{@{}c@{}}Introductory \\ Astronomy\end{tabular}}}}    & \multicolumn{1}{c|}{Q1}         & \multicolumn{1}{c|}{$3.90_{\pm0.54}$}                                                  & \multicolumn{1}{c}{$5.15_{\pm0.27}$} \vrule width 1pt                                               & \multicolumn{1}{c|}{$4.50_{\pm0.16}$}                                               & \multicolumn{1}{c|}{$2.90_{\pm0.61}$}                                                                & \multicolumn{1}{c}{$4.10_{\pm0.61}$} \vrule width 1pt                                                             & \multicolumn{1}{c|}{$4.75_{\pm0.41}$}                                               & \multicolumn{1}{c|}{$4.40_{\pm0.41}$}                                                                & $4.40_{\pm0.48}$                                                             \\ \cmidrule(l){2-10} 
            \multicolumn{1}{c|}{}                         & \multicolumn{1}{c|}{Q2}         & \multicolumn{1}{c|}{$8.20_{\pm0.37}$}                                                  & \multicolumn{1}{c}{$7.55_{\pm0.75}$} \vrule width 1pt                                               & \multicolumn{1}{c|}{$7.60_{\pm0.29}$}                                               & \multicolumn{1}{c|}{$8.30_{\pm0.25}$}                                                                & \multicolumn{1}{c}{$8.40_{\pm0.21}$} \vrule width 1pt                                                             & \multicolumn{1}{c|}{$8.65_{\pm0.20}$}                                               & \multicolumn{1}{c|}{$8.30_{\pm0.28}$}                                                                & $8.50_{\pm0.27}$                                                             \\ \cmidrule(l){2-10} 
            \multicolumn{1}{c|}{}                         & \multicolumn{1}{c|}{Q3}         & \multicolumn{1}{c|}{$7.51_{\pm0.92}$}                                                  & \multicolumn{1}{c}{$7.41_{\pm0.82}$} \vrule width 1pt                                               & \multicolumn{1}{c|}{$6.81_{\pm0.75}$}                                               & \multicolumn{1}{c|}{$7.21_{\pm0.85}$}                                                                & \multicolumn{1}{c}{$6.61_{\pm0.78}$} \vrule width 1pt                                                             & \multicolumn{1}{c|}{$7.61_{\pm0.87}$}                                               & \multicolumn{1}{c|}{$7.31_{\pm0.86}$}                                                                & $7.61_{\pm0.87}$                                                             \\ \cmidrule(l){2-10} 
            \multicolumn{1}{c|}{}                         & \multicolumn{1}{c|}{Q4}         & \multicolumn{1}{c|}{$7.41_{\pm0.86}$}                                                  & \multicolumn{1}{c}{$7.45_{\pm0.46}$} \vrule width 1pt                                               & \multicolumn{1}{c|}{$6.81_{\pm0.78}$}                                               & \multicolumn{1}{c|}{$7.11_{\pm0.91}$}                                                                & \multicolumn{1}{c}{$6.81_{\pm0.94}$} \vrule width 1pt                                                             & \multicolumn{1}{c|}{$7.51_{\pm0.77}$}                                               & \multicolumn{1}{c|}{$6.91_{\pm0.90}$}                                                                & $7.06_{\pm0.89}$                                                             \\ \cmidrule(l){2-10} 
            \multicolumn{1}{c|}{}                         & \multicolumn{1}{c|}{Q5}         & \multicolumn{1}{c|}{$5.51_{\pm0.94}$}                                                  & \multicolumn{1}{c}{$7.40_{\pm0.83}$} \vrule width 1pt                                               & \multicolumn{1}{c|}{$6.41_{\pm0.78}$}                                               & \multicolumn{1}{c|}{$7.40_{\pm0.91}$}                                                                & \multicolumn{1}{c}{$6.41_{\pm1.10}$} \vrule width 1pt                                                             & \multicolumn{1}{c|}{$6.21_{\pm0.90}$}                                               & \multicolumn{1}{c|}{$5.91_{\pm1.06}$}                                                                & $6.36_{\pm1.10}$                                                             \\ \midrule[0.9pt]
            \multicolumn{1}{c|}{\multirow{4}{*}{{\begin{tabular}[c]{@{}c@{}}Astrobiology\end{tabular}}}} & \multicolumn{1}{c|}{Q1}         & \multicolumn{1}{c|}{$6.81_{\pm1.09}$}                                                  & \multicolumn{1}{c}{$7.50_{\pm0.79}$} \vrule width 1pt                                               & \multicolumn{1}{c|}{$7.41_{\pm0.84}$}                                               & \multicolumn{1}{c|}{$6.71_{\pm1.18}$}                                                                & \multicolumn{1}{c}{$6.60_{\pm0.94}$} \vrule width 1pt                                                             & \multicolumn{1}{c|}{$7.50_{\pm0.70}$}                                               & \multicolumn{1}{c|}{$7.11_{\pm0.95}$}                                                                & $7.11_{\pm0.85}$                                                             \\ \cmidrule(l){2-10} 
            \multicolumn{1}{c|}{}                         & \multicolumn{1}{c|}{Q2}         & \multicolumn{1}{c|}{$6.71_{\pm0.83}$}                                                  & \multicolumn{1}{c}{$7.46_{\pm1.02}$} \vrule width 1pt                                               & \multicolumn{1}{c|}{$7.01_{\pm0.76}$}                                               & \multicolumn{1}{c|}{$7.31_{\pm0.83}$}                                                                & \multicolumn{1}{c}{$5.53_{\pm1.30}$} \vrule width 1pt                                                             & \multicolumn{1}{c|}{$7.91_{\pm0.68}$}                                               & \multicolumn{1}{c|}{$7.41_{\pm0.83}$}                                                                & $7.11_{\pm0.98}$                                                             \\ \cmidrule(l){2-10} 
            \multicolumn{1}{c|}{}                         & \multicolumn{1}{c|}{Q3}         & \multicolumn{1}{c|}{$7.89_{\pm0.85}$}                                                  & \multicolumn{1}{c}{$9.04_{\pm0.52}$} \vrule width 1pt                                               & \multicolumn{1}{c|}{$6.70_{\pm0.71}$}                                               & \multicolumn{1}{c|}{$6.49_{\pm0.85}$}                                                                & \multicolumn{1}{c}{$5.02_{\pm1.35}$} \vrule width 1pt                                                             & \multicolumn{1}{c|}{$8.10_{\pm0.33}$}                                               & \multicolumn{1}{c|}{$7.50_{\pm0.41}$}                                                                & $7.50_{\pm0.29}$                                                             \\ \midrule[0.9pt]
            \multicolumn{1}{c|}{\multirow{5.5}{*}{{\begin{tabular}[c]{@{}c@{}}History \& \\ Philosophy of \\ Astronomy \end{tabular}}}}     & \multicolumn{1}{c|}{Q1}         & \multicolumn{1}{c|}{$3.50_{\pm0.21}$}                                                  & \multicolumn{1}{c}{$3.60_{\pm0.21}$} \vrule width 1pt                                               & \multicolumn{1}{c|}{$2.70_{\pm0.14}$}                                               & \multicolumn{1}{c|}{$2.00_{\pm0.00}$}                                                                & \multicolumn{1}{c}{$2.85_{\pm0.20}$} \vrule width 1pt                                                             & \multicolumn{1}{c|}{$3.50_{\pm0.21}$}                                               & \multicolumn{1}{c|}{$3.20_{\pm0.31}$}                                                                & $3.20_{\pm0.22}$                                                             \\ \cmidrule(l){2-10} 
            \multicolumn{1}{c|}{}                         & \multicolumn{1}{c|}{Q2}         & \multicolumn{1}{c|}{$2.39_{\pm0.29}$}                                                  & \multicolumn{1}{c}{$3.69_{\pm0.20}$} \vrule width 1pt                                               & \multicolumn{1}{c|}{$2.90_{\pm0.22}$}                                               & \multicolumn{1}{c|}{$1.80_{\pm0.31}$}                                                                & \multicolumn{1}{c}{$2.56_{\pm0.25}$} \vrule width 1pt                                                             & \multicolumn{1}{c|}{$3.25_{\pm0.23}$}                                               & \multicolumn{1}{c|}{$3.10_{\pm0.17}$}                                                                & $2.95_{\pm0.23}$                                                             \\ \cmidrule(l){2-10} 
            \multicolumn{1}{c|}{}                         & \multicolumn{1}{c|}{Q3}         & \multicolumn{1}{c|}{$2.70_{\pm0.20}$}                                                  & \multicolumn{1}{c}{$3.40_{\pm0.38}$} \vrule width 1pt                                               & \multicolumn{1}{c|}{$2.69_{\pm0.32}$}                                               & \multicolumn{1}{c|}{$1.20_{\pm0.42}$}                                                                & \multicolumn{1}{c}{$1.70_{\pm0.38}$} \vrule width 1pt                                                             & \multicolumn{1}{c|}{$3.65_{\pm0.14}$}                                               & \multicolumn{1}{c|}{$3.20_{\pm0.19}$}                                                                & $3.20_{\pm0.18}$                                                             \\ \cmidrule(l){2-10} 
            \multicolumn{1}{c|}{}                         & \multicolumn{1}{c|}{Q4}         & \multicolumn{1}{c|}{$2.20_{\pm0.28}$}                                                  & \multicolumn{1}{c}{$3.80_{\pm0.19}$} \vrule width 1pt                                               & \multicolumn{1}{c|}{$3.00_{\pm0.14}$}                                               & \multicolumn{1}{c|}{$1.10_{\pm0.33}$}                                                                & \multicolumn{1}{c}{$2.20_{\pm0.19}$} \vrule width 1pt                                                             & \multicolumn{1}{c|}{$3.25_{\pm0.24}$}                                               & \multicolumn{1}{c|}{$2.70_{\pm0.28}$}                                                                & $2.95_{\pm0.23}$                                                             \\ \bottomrule
        \end{tabular}
    \end{adjustbox}
\end{table*}

\section{Experimental Setup}
\label{section:experimental-setup}


\textbf{Data.}
This study employs data from three MOOCs: Astrobiology \citep{impey_astrobiology}, Introductory Astronomy \citep{impey_astronomy}, and the History and Philosophy of Astronomy \citep{impey_history}.
The process of obtaining the data followed a strict ethical roadmap, guided by an Institutional Review Board from the university.
Participants voluntarily completed a survey and provided formal consent for their data to be used in the research. Also, all identifying information was anonymized. Since the original dataset had most assignments clustered at either very high or low grades, we sampled data to cover a wider range of grades for a more comprehensive evaluation of LLM grading performance.


Most students in these MOOCs are between the ages of 25 and 44. A slight majority of students are male across all courses, with the smallest gender gap in the Astrobiology course (45\% female vs. 54\% male) and the largest in the History and Philosophy course (34\% female vs. 65\% male). Geographically, most students are from the United States and India.

Regarding the attributes of the assignments, students are advised to limit their answers to 500 words, although this restriction is not strictly enforced. Specifically, in the Introductory Astronomy course, there are five assignments, each worth 9 points, except for the first question, which is valued at 6 points. The Astrobiology course comprises three questions, each worth 10 points, while the Philosophy and History of Astronomy course includes four questions, each valued at 4 points.\footnote{See Appendix \ref{appendix:details_of_questions} for comprehensive details on questions.}





Following previous studies \citep[inter alia]{pankiewicz2023large,10343457,gabbay2024combining}, we use a subset of data for our experiments.
Specifically, we analyze 12 assignments from a group of 10 students in each of the three courses, resulting in analyzing 120 questions per prompt.
Note that the data in this study is sourced from a proprietary repository, proactively accounting for the potential issue of data contamination in LLMs \cite{DBLP:journals/corr/abs-2308-08493,DBLP:journals/corr/abs-2311-06233}.

\vspace{0.08cm}

\noindent\textbf{Setting.}
For all experiments with GPT-4 and GPT-3.5, we employ the \texttt{gpt-4-0613} and \texttt{gpt-3.5-turbo-0613} snapshots, accessed via the OpenAI API. We set the temperature to zero and the maximum token limit to 2048 to prompt LLMs. These hyperparameters remain consistent across all settings throughout our evaluations.

\vspace{0.08cm}

\noindent\textbf{Instructor Grading.} 
In our study, we regard the grades given by instructors (university professors) as our ground truth. These professors are experts in the relevant fields of each course and carefully assess the assignments in question. The grading process is double-blind, i.e., instructors are unaware of peer grading and students' information.




\vspace{0.08cm}

\noindent\textbf{Peer Grading.} 
Each writing assignment is evaluated by three to four randomly selected classmates using grading rubrics provided by the instructor, identical to the one used for instructor grading. Despite anonymous grading to maintain impartiality, the final grade for each assignment is determined by calculating the median of scores from peer graders.

\vspace{0.08cm}

\noindent\textbf{Bootstrap Resampling.}
In all settings involving bootstrap resampling, we conduct our experiments with 10,000 iterations using replacement. Additionally, in these settings, the significance of the results is reported based on a $p$-value threshold of 0.05 where relevant.

\section{Results and Discussion}

Tables \ref{tab:raw-resullts} and \ref{tab:bootstrap-results} present the outcomes of our assessment of writing assignments using GPT-4 and GPT-3.5 across three MOOC subjects. These assessments are grounded on three different prompts devised in our study. Table \ref{tab:raw-resullts} shows the average grades derived directly from grading the assignments of 10 students per question for each course, whereas Table \ref{tab:bootstrap-results} displays the results after conducting bootstrap resampling with 10,000 iterations on the data from Table \ref{tab:raw-resullts}. 
In the following, we provide both our qualitative and quantitative analyses.

\subsection{Qualitative Analysis: Average Grades Evaluation}

Qualitatively, our results unfold several insights:

\vspace{0.08cm}

\noindent \textbf{(1)} First and foremost, Table \ref{tab:raw-resullts} reports no statistically significant differences in average grades produced by LLMs versus instructor-assigned grades. This suggests that LLM grading, akin to peer grading, is capable of generating grades consistent with those assigned by instructors.

\vspace{0.08cm}

\noindent \textbf{(2)} Tables \ref{tab:raw-resullts} and \ref{tab:bootstrap-results} reveal that GPT-4 overall outperforms GPT-3.5 by generating grades that are more closely aligned with the instructor-assigned grades, especially in the context of the History and Philosophy of Astronomy. 

\vspace{0.08cm}

\noindent \textbf{(3)} Focusing on the prompts in Tables \ref{tab:raw-resullts} and \ref{tab:bootstrap-results}, ones that integrate both instructor-provided rubrics and LLM-generated rubrics give rise to grades that are better aligned with those from the instructors. This is true for both GPT-4 and GPT-3.5.

\vspace{0.08cm}

\noindent \textbf{(4)} Looking at Tables \ref{tab:raw-resullts} and \ref{tab:bootstrap-results}, in almost all scenarios, GPT-4, when prompted with instructor-provided correct answers and rubrics, generates grades that are either superior to or on par with peer grades. This even includes the History and Philosophy of Astronomy, which needs speculative or imaginative thinking abilities. 

\vspace{0.08cm}

\noindent \textbf{(5)} When GPT-4 serves as the underlying base model, LLM-generated rubrics demonstrate comparable performance to instructor-provided rubrics, as illustrated in Tables \ref{tab:raw-resullts} and \ref{tab:bootstrap-results}. This finding is interesting, indicating the potential use of LLMs for rubric generation in educational contexts, enabling the automation of grading processes with only the input of correct answers.

\vspace{0.08cm}

\noindent \textbf{(6)} Tables \ref{tab:mae-org} and \ref{tab:mae-bootstrap} (in Appendix \ref{appendix:mae-results}) show that LLMs more often produce the lowest MAEs across all questions compared to peer grading, indicating that LLM-assigned grades are more accurate, with fewer errors or deviations from the instructor-assigned grades. Additionally, GPT-4 consistently assigns grades with lower MAEs than GPT-3.5.

\vspace{0.08cm}

\noindent \textbf{(7)} From the results listed in Tables \ref{tab:raw-resullts} and \ref{tab:bootstrap-results}, it is clear that the most difficult course to grade is the History and Philosophy of Astronomy. Both LLMs and peer graders struggle to match instructor-assigned grades closely. Nevertheless, even in these cases, GPT-4 outperforms peer grading, generating grades more aligned with instructor-assigned ones.


\subsection{Quantitative Analysis: Question-by-Question Evaluation}

Quantitatively, after assessing grades and feedback produced by LLMs on a question-by-question basis, we summarize our main observations as follows:

\vspace{0.08cm}

\noindent \textbf{(1)} Generally, LLM-assigned grades are higher and more generous than instructor-assigned grades. Although these scores are higher, they align more closely with instructor-assigned grades than peer grades do, particularly in the History and Philosophy of Astronomy. Despite this alignment, some instability was observed. For example, in the Astrobiology course, both models vary dramatically regardless of the prompt strategy employed, with GPT-3.5 scoring low and GPT-4 scoring high, differing from the instructor-assigned scores. We also observed that LLMs encounter challenges based on the length of students' answers, particularly with extremely short or excessively long responses.
In such cases, while GPT-4 handled short answers well, it struggled with long ones.

\vspace{0.08cm}

\noindent \textbf{(2)} There is a bias towards middle scores and against extremes, especially when ZCoT combines with correct answers and, to a lesser extent, when ZCoT couples with correct answers and LLM-generated rubrics. Both prompts lead GPT-3.5 and GPT-4 to be less likely to assign full marks compared to the instructor. Also, it is rare for LLMs to give a zero score on the same questions as the instructor. This is particularly evident in the History and Philosophy of Astronomy, where GPT-3.5 rarely gave full credit and assigned more zeros than the instructor for the ZCoT with correct answers and LLM-generated rubrics prompt, which performed well with other assignments and with GPT-4 on the same assignments.

\vspace{0.08cm}

\noindent \textbf{(3)} One of the most interesting results of this study is the feedback from LLMs. Both GPT-3.5 and GPT-4 produced subject-appropriate explanations specific enough to justify point deductions, referring to rubric criteria when provided. Most importantly, LLM feedback was more consistent and comprehensive than peer graders, who often failed to leave feedback or provided insufficient detail for students to improve. In contrast, LLMs, regardless of the prompt strategy, provided accurate and constructive feedback. Namely, when GPT-4 employs ZCoT with correct answers and instructor-provided rubrics, it often matches instructor feedback, including details leading to point deductions.

\vspace{0.08cm}

\noindent \textbf{(4)} In our experiments with GPT-3.5 and GPT-4, we did not observe any instances of hallucinations when generating grades and feedback for assignments. This can be attributed to the inclusion of detailed rubrics and correct answers in the input prompt, which guide the output into a more structured and factual format. These findings suggest the potential of LLMs to serve as reliable automated grading systems when provided with sufficient guidance in the input prompt.

\section{Conclusion}
We studied the feasibility and validity of using large language models (LLMs) to replace peer grading in massive open online courses (MOOCs). To this end, we developed three distinct prompts based on the zero-shot chain-of-thought (ZCoT) prompting technique, each with varying in-context information. Specifically, our methods include ZCoT with instructor-provided correct answers, ZCoT with instructor-provided correct answers and rubrics, as well as ZCoT with instructor-provided correct answers and LLM-generated rubrics.

We evaluated the effectiveness of each prompt using two LLMs, GPT-4 and GPT-3.5, by grading assignments from three MOOCs: Introductory Astronomy, Astrobiology, and the History and Philosophy of Astronomy. Our findings indicated that ZCoT with instructor-provided correct answers and rubrics, particularly with GPT-4, consistently provides grades that are more aligned with instructor-assigned grades compared to peer grading. Further, we found that grading courses requiring reasoning/imaginative skills, e.g., the History and Philosophy of Astronomy, is more challenging. In such cases, GPT-4 also outperforms peer grading.

Overall, our results show the potential of using LLMs in MOOCs to automate grading systems and replace peer grading, thus enhancing the learning experience for millions of online students globally.



\section{Limitations}
In this research, we studied the promise of LLM grading as a substitute for peer grading in MOOCs to automate the grading process. Yet, opportunities exist for refining the alignment between LLM-assigned and instructor-assigned grades, notably in advanced disciplines such as Philosophy and Mathematics, which demand robust reasoning abilities. Hence, further research is warranted to refine grading methodologies for improved congruence with instructor evaluations.

\section{Ethical Considerations}

The use of LLMs for automated grading introduces several ethical concerns that warrant careful examination. One major concern is fairness in grading. In fact, biases present in the training data of LLMs could lead to systematic disparities in scores across different demographic groups. For example, if an LLM has been trained on data that reflects societal biases, it might inadvertently favor certain linguistic styles, cultural norms, argumentation styles, or perspectives, thereby disadvantaging students who do not conform to these patterns.

Students' perceptions of machine-generated feedback also present ethical concerns. Feedback from LLMs might be perceived as less personal or less tailored to individual learning paths compared to human-generated feedback. It is essential to communicate clearly with students about how automated grading systems using LLMs operate, the rationale behind their feedback, and the ways in which human oversight is incorporated into the grading process.

In conclusion, while automated grading with LLMs offers significant benefits in efficiency and scalability, it is imperative to address the aforesaid ethical considerations proactively. Ensuring fairness, minimizing biases, and fostering positive perceptions among students are crucial steps toward leveraging the advantages of LLMs while upholding the integrity of the educational experience.

\bibliography{custom}

\begin{thebibliography}{41}
\providecommand{\natexlab}[1]{#1}

\bibitem[{Alseddiqi et~al.(2023)Alseddiqi, AL-Mofleh, Albalooshi, and Najam}]{Alseddiqi_2023}
Mohamed Alseddiqi, Anwar AL-Mofleh, Leena Albalooshi, and Osama Najam. 2023.
\newblock \href {https://doi.org/10.24018/ejedu.2023.4.4.686} {Revolutionizing online learning: The potential of chatgpt in massive open online courses}.
\newblock \emph{European Journal of Education and Pedagogy}.

\bibitem[{Chang and Ginter(2024)}]{chang2024automatic}
Li-Hsin Chang and Filip Ginter. 2024.
\newblock Automatic short answer grading for finnish with chatgpt.
\newblock In \emph{Proceedings of the AAAI Conference on Artificial Intelligence}, volume~38, pages 23173--23181.

\bibitem[{Del Carpio~Gutierrez et~al.(2024)Del Carpio~Gutierrez, Denny, and Luxton-Reilly}]{10.1145/3626252.3630863}
Andre Del Carpio~Gutierrez, Paul Denny, and Andrew Luxton-Reilly. 2024.
\newblock \href {https://doi.org/10.1145/3626252.3630863} {Evaluating automatically generated contextualised programming exercises}.
\newblock In \emph{Proceedings of the 55th ACM Technical Symposium on Computer Science Education V. 1}, SIGCSE 2024, page 289–295, New York, NY, USA. Association for Computing Machinery.

\bibitem[{Devlin et~al.(2018)Devlin, Chang, Lee, and Toutanova}]{devlin2018bert}
Jacob Devlin, Ming-Wei Chang, Kenton Lee, and Kristina Toutanova. 2018.
\newblock Bert: Pre-training of deep bidirectional transformers for language understanding.
\newblock \emph{arXiv preprint arXiv:1810.04805}.

\bibitem[{Efron(1979)}]{10.1214/aos/1176344552}
B.~Efron. 1979.
\newblock \href {https://doi.org/10.1214/aos/1176344552} {{Bootstrap Methods: Another Look at the Jackknife}}.
\newblock \emph{The Annals of Statistics}, 7(1):1 -- 26.

\bibitem[{Efron and Tibshirani(1993)}]{EfroTibs93}
Bradley Efron and Robert~J. Tibshirani. 1993.
\newblock \emph{An Introduction to the Bootstrap}.
\newblock Number~57 in Monographs on Statistics and Applied Probability. Chapman \& Hall/CRC, Boca Raton, Florida, USA.

\bibitem[{Formanek et~al.(2017)Formanek, Wenger, Buxner, Impey, and Sonam}]{FORMANEK2017243}
Martin Formanek, Matthew~C. Wenger, Sanlyn~R. Buxner, Chris~D. Impey, and Tenzin Sonam. 2017.
\newblock \href {https://doi.org/10.1016/j.compedu.2017.05.019} {Insights about large-scale online peer assessment from an analysis of an astronomy mooc}.
\newblock \emph{Computers \& Education}, 113:243--262.

\bibitem[{Funayama et~al.(2022)Funayama, Sato, Matsubayashi, Mizumoto, Suzuki, and Inui}]{Hiroaki_2022}
Hiroaki Funayama, Tasuku Sato, Yuichiroh Matsubayashi, Tomoya Mizumoto, Jun Suzuki, and Kentaro Inui. 2022.
\newblock \href {https://doi.org/10.48550/arxiv.2206.08288} {Balancing cost and quality: An exploration of human-in-the-loop frameworks for automated short answer scoring}.
\newblock \emph{Cornell University - arXiv}.

\bibitem[{Gabbay and Cohen(2024)}]{gabbay2024combining}
Hagit Gabbay and Anat Cohen. 2024.
\newblock Combining llm-generated and test-based feedback in a mooc for programming.
\newblock In \emph{Proceedings of the Eleventh ACM Conference on Learning@ Scale}, pages 177--187.

\bibitem[{Gamage et~al.(2021)Gamage, Gamage, Staubitz, Staubitz, Whiting, and Whiting}]{Gamage_2021}
Dilrukshi Gamage, Dilrukshi Gamage, Thomas Staubitz, Thomas Staubitz, Mark~E. Whiting, and Mark~E. Whiting. 2021.
\newblock \href {https://doi.org/10.1080/01587919.2021.1911626} {Peer assessment in moocs: Systematic literature review}.
\newblock \emph{Distance Education}.

\bibitem[{Golchin and Surdeanu(2023{\natexlab{a}})}]{DBLP:journals/corr/abs-2311-06233}
Shahriar Golchin and Mihai Surdeanu. 2023{\natexlab{a}}.
\newblock \href {https://doi.org/10.48550/ARXIV.2311.06233} {Data contamination quiz: {A} tool to detect and estimate contamination in large language models}.
\newblock \emph{CoRR}, abs/2311.06233.

\bibitem[{Golchin and Surdeanu(2023{\natexlab{b}})}]{DBLP:journals/corr/abs-2308-08493}
Shahriar Golchin and Mihai Surdeanu. 2023{\natexlab{b}}.
\newblock \href {https://doi.org/10.48550/ARXIV.2308.08493} {Time travel in llms: Tracing data contamination in large language models}.
\newblock \emph{CoRR}, abs/2308.08493.

\bibitem[{Heickal and Lan(2024)}]{heickal2024generating}
Hasnain Heickal and Andrew Lan. 2024.
\newblock Generating feedback-ladders for logical errors in programming using large language models.
\newblock \emph{arXiv preprint arXiv:2405.00302}.

\bibitem[{Impey(2024{\natexlab{a}})}]{impey_astrobiology}
Chris Impey. 2024{\natexlab{a}}.
\newblock \href {https://www.coursera.org/learn/astrobiology-exploring-other-worlds} {Astrobiology: Exploring other worlds}.
\newblock Accessed: 2024-06-08.

\bibitem[{Impey(2024{\natexlab{b}})}]{impey_astronomy}
Chris Impey. 2024{\natexlab{b}}.
\newblock \href {https://www.coursera.org/learn/astro} {Astronomy: Exploring time and space}.
\newblock Accessed: 2024-06-08.

\bibitem[{Impey(2024{\natexlab{c}})}]{impey_history}
Chris Impey. 2024{\natexlab{c}}.
\newblock \href {https://www.coursera.org/learn/knowing-the-universe} {Knowing the universe: An introduction to astronomy}.
\newblock Accessed: 2024-06-08.

\bibitem[{Impey et~al.(2016)Impey, Wenger, Formanek, and Buxner}]{impey2016bringing}
Chris Impey, Matthew Wenger, Martin Formanek, and Sanlyn Buxner. 2016.
\newblock Bringing the universe to the world: lessons learned from a massive open online class on astronomy.
\newblock \emph{Communicating Astronomy with the Public}, 21:20--30.

\bibitem[{Impey et~al.(2015)Impey, Wenger, and Austin}]{Impey_Wenger_Austin_2015}
Chris~D. Impey, Matthew~C. Wenger, and Carmen~L. Austin. 2015.
\newblock \href {https://doi.org/10.19173/irrodl.v16i1.1983} {Astronomy for astronomical numbers: A worldwide massive open online class}.
\newblock \emph{The International Review of Research in Open and Distributed Learning}, 16(1).

\bibitem[{Jacobs and Jaschke(2024)}]{jacobs2024evaluating}
Sven Jacobs and Steffen Jaschke. 2024.
\newblock Evaluating the application of large language models to generate feedback in programming education.
\newblock \emph{arXiv preprint arXiv:2403.09744}.

\bibitem[{Kasneci et~al.(2023)Kasneci, Sessler, Küchemann, Bannert, Dementieva, Fischer, Gasser, Groh, Günnemann, Hüllermeier, Krusche, Kutyniok, Michaeli, Nerdel, Pfeffer, Poquet, Sailer, Schmidt, Seidel, Stadler, Weller, Kühn, and Kasneci}]{Kasneci_2023}
Enkelejda Kasneci, Kathrin Sessler, Stefan Küchemann, Maria Bannert, Daryna Dementieva, Frank Fischer, Urs Gasser, Georg Groh, Stephan Günnemann, Eyke Hüllermeier, Stepha Krusche, Gitta Kutyniok, Tilman Michaeli, Claudia Nerdel, Jürgen Pfeffer, Oleksandra Poquet, Michael Sailer, Albrecht Schmidt, Tina Seidel, Matthias Stadler, J.~Weller, Jochen Kühn, and Gjergji Kasneci. 2023.
\newblock \href {https://doi.org/10.1016/j.lindif.2023.102274} {Chatgpt for good? on opportunities and challenges of large language models for education}.
\newblock \emph{Learning and Individual Differences}.

\bibitem[{Kiesler et~al.(2023)Kiesler, Lohr, and Keuning}]{10343457}
Natalie Kiesler, Dominic Lohr, and Hieke Keuning. 2023.
\newblock \href {https://doi.org/10.1109/FIE58773.2023.10343457} {Exploring the potential of large language models to generate formative programming feedback}.
\newblock In \emph{2023 IEEE Frontiers in Education Conference (FIE)}, pages 1--5.

\bibitem[{Kocmi and Federmann(2023)}]{kocmi-federmann-2023-large}
Tom Kocmi and Christian Federmann. 2023.
\newblock \href {https://aclanthology.org/2023.eamt-1.19} {Large language models are state-of-the-art evaluators of translation quality}.
\newblock In \emph{Proceedings of the 24th Annual Conference of the European Association for Machine Translation}, pages 193--203, Tampere, Finland. European Association for Machine Translation.

\bibitem[{Kojima et~al.(2022)Kojima, Gu, Reid, Matsuo, and Iwasawa}]{NEURIPS2022_8bb0d291}
Takeshi Kojima, Shixiang~(Shane) Gu, Machel Reid, Yutaka Matsuo, and Yusuke Iwasawa. 2022.
\newblock Large language models are zero-shot reasoners.
\newblock In \emph{Advances in Neural Information Processing Systems}, volume~35, pages 22199--22213.

\bibitem[{Liu et~al.(2023)Liu, Iter, Xu, Wang, Xu, and Zhu}]{liu2023g}
Yang Liu, Dan Iter, Yichong Xu, Shuohang Wang, Ruochen Xu, and Chenguang Zhu. 2023.
\newblock G-eval: Nlg evaluation using gpt-4 with better human alignment.
\newblock \emph{arXiv preprint arXiv:2303.16634}.

\bibitem[{Luo et~al.(2023)Luo, Xie, and Ananiadou}]{luo2023chatgpt}
Zheheng Luo, Qianqian Xie, and Sophia Ananiadou. 2023.
\newblock Chatgpt as a factual inconsistency evaluator for text summarization.
\newblock \emph{arXiv preprint arXiv:2303.15621}.

\bibitem[{Matelsky et~al.(2023)Matelsky, Parodi, Liu, Lange, and Kording}]{matelsky2023large}
Jordan~K Matelsky, Felipe Parodi, Tony Liu, Richard~D Lange, and Konrad~P Kording. 2023.
\newblock A large language model-assisted education tool to provide feedback on open-ended responses.
\newblock \emph{arXiv preprint arXiv:2308.02439}.

\bibitem[{Meyer et~al.(2024)Meyer, Jansen, Schiller, Liebenow, Steinbach, Horbach, and Fleckenstein}]{meyer2024using}
Jennifer Meyer, Thorben Jansen, Ronja Schiller, Lucas~W Liebenow, Marlene Steinbach, Andrea Horbach, and Johanna Fleckenstein. 2024.
\newblock Using llms to bring evidence-based feedback into the classroom: Ai-generated feedback increases secondary students’ text revision, motivation, and positive emotions.
\newblock \emph{Computers and Education: Artificial Intelligence}, 6:100199.

\bibitem[{Morris et~al.(2023)Morris, Crossley, Holmes, and Trumbore}]{Morris_2023}
Wesley Morris, Scott~A. Crossley, Langdon Holmes, and Anne Trumbore. 2023.
\newblock \href {https://doi.org/10.1145/3576050.3576098} {Using transformer language models to validate peer-assigned essay scores in massive open online courses (moocs)}.
\newblock \emph{International Conference on Learning Analytics and Knowledge}.

\bibitem[{OpenAI(2023)}]{DBLP:journals/corr/abs-2303-08774}
OpenAI. 2023.
\newblock \href {https://doi.org/10.48550/ARXIV.2303.08774} {{GPT-4} technical report}.
\newblock \emph{CoRR}, abs/2303.08774.

\bibitem[{OpenAI(2024)}]{khan_openai}
OpenAI. 2024.
\newblock \href {https://openai.com/index/khan-academy/} {Openai partners with khan academy to offer math exercises}.
\newblock Accessed: 2024-06-08.

\bibitem[{Ouyang et~al.(2022)Ouyang, Wu, Jiang, Almeida, Wainwright, Mishkin, Zhang, Agarwal, Slama, Ray, Schulman, Hilton, Kelton, Miller, Simens, Askell, Welinder, Christiano, Leike, and Lowe}]{DBLP:conf/nips/Ouyang0JAWMZASR22}
Long Ouyang, Jeffrey Wu, Xu~Jiang, Diogo Almeida, Carroll~L. Wainwright, Pamela Mishkin, Chong Zhang, Sandhini Agarwal, Katarina Slama, Alex Ray, John Schulman, Jacob Hilton, Fraser Kelton, Luke Miller, Maddie Simens, Amanda Askell, Peter Welinder, Paul~F. Christiano, Jan Leike, and Ryan Lowe. 2022.
\newblock \href {http://papers.nips.cc/paper\_files/paper/2022/hash/b1efde53be364a73914f58805a001731-Abstract-Conference.html} {Training language models to follow instructions with human feedback}.
\newblock In \emph{NeurIPS}.

\bibitem[{Pankiewicz and Baker(2023)}]{pankiewicz2023large}
Maciej Pankiewicz and Ryan~S Baker. 2023.
\newblock Large language models (gpt) for automating feedback on programming assignments.
\newblock \emph{arXiv preprint arXiv:2307.00150}.

\bibitem[{Pinto et~al.(2023)Pinto, Cardoso-Pereira, Ribeiro, Lucena, Souza, and Gama}]{Pinto_2023}
Gustavo Pinto, Isadora Cardoso-Pereira, Danilo~Braga Ribeiro, Danilo Lucena, Alberto Souza, and Kiev Gama. 2023.
\newblock \href {https://doi.org/10.1145/3613372.3614197} {Large language models for education: Grading open-ended questions using chatgpt}.
\newblock \emph{Brazilian Symposium on Software Engineering}.

\bibitem[{Riordan et~al.(2017)Riordan, Riordan, Horbach, Horbach, Cahill, Cahill, Zesch, Zesch, Lee, and Lee}]{Riordan_2017}
Brian Riordan, Brian Riordan, Andrea Horbach, Andrea Horbach, Aoife Cahill, Aoife Cahill, Torsten Zesch, Torsten Zesch, Chong~Min Lee, and Chong~Min Lee. 2017.
\newblock \href {https://doi.org/10.18653/v1/w17-5017} {Investigating neural architectures for short answer scoring}.
\newblock \emph{BEA@EMNLP}.

\bibitem[{Stahl et~al.(2024)Stahl, Biermann, Nehring, and Wachsmuth}]{stahl2024exploring}
Maja Stahl, Leon Biermann, Andreas Nehring, and Henning Wachsmuth. 2024.
\newblock Exploring llm prompting strategies for joint essay scoring and feedback generation.
\newblock \emph{arXiv preprint arXiv:2404.15845}.

\bibitem[{Tripaldelli et~al.(2024)Tripaldelli, Pozek, and Butka}]{10578705}
Alessia Tripaldelli, George Pozek, and Brian Butka. 2024.
\newblock \href {https://doi.org/10.1109/EDUCON60312.2024.10578705} {Leveraging prompt engineering on chatgpt for enhanced learning in cec330: Digital system design in aerospace}.
\newblock In \emph{2024 IEEE Global Engineering Education Conference (EDUCON)}, pages 1--9.

\bibitem[{Vaswani et~al.(2017)Vaswani, Shazeer, Parmar, Uszkoreit, Jones, Gomez, Kaiser, and Polosukhin}]{vaswani2017attention}
Ashish Vaswani, Noam Shazeer, Niki Parmar, Jakob Uszkoreit, Llion Jones, Aidan~N Gomez, {\L}ukasz Kaiser, and Illia Polosukhin. 2017.
\newblock Attention is all you need.
\newblock \emph{Advances in neural information processing systems}, 30.

\bibitem[{Zeng et~al.(2023)Zeng, Yu, Gao, Meng, Goyal, and Chen}]{zeng2023evaluating}
Zhiyuan Zeng, Jiatong Yu, Tianyu Gao, Yu~Meng, Tanya Goyal, and Danqi Chen. 2023.
\newblock Evaluating large language models at evaluating instruction following.
\newblock \emph{arXiv preprint arXiv:2310.07641}.

\bibitem[{Zhang et~al.(2024)Zhang, Dong, Shi, Price, Matsuda, and Xu}]{zhang2024students}
Zhengdong Zhang, Zihan Dong, Yang Shi, Thomas Price, Noboru Matsuda, and Dongkuan Xu. 2024.
\newblock Students’ perceptions and preferences of generative artificial intelligence feedback for programming.
\newblock In \emph{Proceedings of the AAAI Conference on Artificial Intelligence}, volume~38, pages 23250--23258.

\bibitem[{Zhu et~al.(2023)Zhu, Wang, and Wang}]{zhu2023judgelm}
Lianghui Zhu, Xinggang Wang, and Xinlong Wang. 2023.
\newblock Judgelm: Fine-tuned large language models are scalable judges.
\newblock \emph{arXiv preprint arXiv:2310.17631}.

\bibitem[{Zhuo(2023)}]{zhuo2023ice}
Terry~Yue Zhuo. 2023.
\newblock Ice-score: Instructing large language models to evaluate code.
\newblock \emph{arXiv preprint arXiv:2304.14317}.

\end{thebibliography}

\appendix

\section{Details on Assignment Questions}
\label{appendix:details_of_questions}

To offer comprehensive insight into the question structure of each course, we list all questions per course below. Note that we cannot publicly release all correct answers and rubrics to these questions due to the active status of the respective courses on Coursera.
The questions are as follows:

\begin{itemize}
    \item \textbf{Course:} Introductory Astronomy
    \begin{enumerate}
        \item In terms of the scientific method, how does astronomy differ from a lab science like 
        chemistry or biology? 
        How can astronomers be confident in their understanding of 
        objects that are remote from the Earth?
        Ancient cultures built some impressive structures that incorporated astronomical 
        functions and information (Stonehenge, Chichen Itza, the Great Pyramid). A friend or 
        acquaintance of yours tries to argue that some of these structures and artifacts are 
        evidence of ``ancient astronauts'' or visits by intelligent aliens. How would you 
        rebut or argue against this idea?

        \item What are the advantages of large telescopes? Provide at least one.
        Why do astronomers want telescopes in space when putting them there is expensive?
        What are some examples of wavelength regions beyond the spectrum of visible light 
        where astronomers can learn about the universe? Provide at least two.

        \item What are the two main, indirect methods for finding exoplanets?
        Why is it so difficult to see exoplanets directly in an image?
        What are some similarities or differences between our Solar System and new, distant 
        planet systems? Provide at least one similarity and/or difference.

        \item What is the source or cause of the Sun’s light, and how do all the elements in the periodic table get produced?
        What is the general process by which a large diffuse cloud of gas turns into a star 
        and surrounding planets?
        Name the two end states of stars much more massive than the Sun and describe 
        their physical properties?

        \item Why do astronomers often say that large telescopes are like time machines, or 
        equivalently, why is distant light old light?
        What is the evidence that the universe began in a hot, dense state of 13.8 billion 
        years ago?
        The atoms in our bodies and in all the stars in all 100 galaxies form a small 
        percentage of the contents of the universe. What are the two dominant ingredients of 
        the universe and why are astronomers so unsure of their physical nature?
    \end{enumerate}
\end{itemize}

\begin{itemize}
    \item \textbf{Course:} Astrobiology
    \begin{enumerate}
        \item Clearly identify the detection methods used to gather data for each exoplanet.  Briefly explain how each detection method works. Correctly identify both detection methods. Clearly explain how each detection method works. Correctly identify which physical characteristics can be learned from each set of data, and explain why. Clearly identify the physical characteristics of both exoplanets. Clearly identify one exoplanet as Earth-like.

        \item Discuss how habitable zone range and spectral type are related.
        
        Yousef says that all three Earth-like planets likely have liquid surface water because they all orbit at 1 astronomical unit (AU).  Since the Earth orbits at 1 AU from the sun, and we know Earth has liquid surface water, then these exoplanets should as well. Address Yousef's statement. Clearly state whether you agree or disagree with the conclusion. Explain your answer with evidence and use data to support your answer.
        
        Lora says that both exoplanets 2 and 3 will have liquid water, but not exoplanet 1.  The star for exoplanet 1 is spectral type A, which is too big and hot and would evaporate water on exoplanets. But exoplanet 2 and exoplanet 3 orbit around favorable spectral types G and M, therefore they likely have liquid surface water.
        Clearly state whether you agree or disagree with Lora's conclusion. Explain your answer with evidence and use data to support your answer.

        \item Discuss why/whether it is plausible that exobiology exists at all, in terms of exoplanet type, spectral type, and orbital distance. 
        Clearly state the geologic eon and make a strong argument for the state of exobiology, given their choice of geologic eon. Discusses whether the exobiology is unicellular, multicellular, intelligent, etc. Use geologic eon or age, exoplanet type, and examples from class to support the argument. Present arguments in a clear, logical way that demonstrates understanding of concepts. Correctly use scientific terms or language.  Connections between concepts should be well developed. 
        
    \end{enumerate}
\end{itemize}

\begin{itemize}
    \item \textbf{Course:} History \& Philosophy of Astronomy
    \begin{enumerate}
        \item  Describe the practical and philosophical importance of astronomy for humans living a nomadic lifestyle in 20,000 B.C. This assignment will be graded on the clarity and completeness of your answer as well as your use of course topics to support your claims. Your answer should be between 200 and 300 words (about one paragraph).

        \item According to the current understanding of the universe, the cosmos has a definite beginning but an infinite future. What are the philosophical problems and implications of this? This assignment will be graded on the clarity and completeness of your answer as well as your use of course topics to support your claims. Your answer should be between 200 and 300 words (about one paragraph).
        
        \item According to Enlightenment philosophy, why might liberty and personal rights be connected to the pursuit of science? This assignment will be graded on the clarity and completeness of your answer as well as your use of course topics to support your claims. Your answer should be between 200 and 300 words (about one paragraph). 
        
        \item  Imagine if the universe was eternal and unchanging as proposed by the steady state model in the 1950’s. What are the philosophical and scientific implications of this? This assignment will be graded on the clarity and completeness of your answer as well as your use of course topics to support your claims. Your answer should be between 200 and 300 words (about one paragraph). 
        
    \end{enumerate}
\end{itemize}

\section{Mean Absolute Error Results}
\label{appendix:mae-results}

Tables \ref{tab:mae-org} and \ref{tab:mae-bootstrap} present the MAE values for LLM-assigned grades and peer grades relative to instructor-assigned grades. The results show that LLMs generally achieve lower MAE values than peer grades, with GPT-4 most frequently obtaining the lowest MAEs.

\begin{table*}[!]
    \centering
    \caption{MAE values for 10 students' writing assignments, comparing LLM-assigned and peer grades to instructor-assigned grades. \textbf{Bold} values indicate the lowest MAE for each question.}
    \label{tab:mae-org}
    \footnotesize

    \begin{adjustbox}{width=\textwidth,center}

        \begin{tabular}{@{}cccccccccc@{}}
            \toprule
                                                                                         &                                                                                        & \multicolumn{1}{l}{}                                                                & \multicolumn{3}{c}{\textbf{GPT-3.5}}                                                                                                                                                                                                                  & \multicolumn{3}{c}{\textbf{GPT-4}}                                                                                                                                                                                               \\ \cmidrule(lr){4-6}  \cmidrule(l){7-9}
            \multicolumn{1}{c|}{\textbf{Courses}}         & \multicolumn{1}{c|}{\textbf{\raisebox{-2.5ex}{\rotatebox{90}{Questions}}}}  & \multicolumn{1}{c}{\textbf{\begin{tabular}[c]{@{}c@{}}Peer\\ Grades\end{tabular}}} \vrule width 1pt & \multicolumn{1}{c|}{\textbf{\begin{tabular}[c]{@{}c@{}}ZCoT W/\\ Answers\end{tabular}}} & \multicolumn{1}{c|}{\textbf{\begin{tabular}[c]{@{}c@{}}ZCoT W/\\ Answers \&\\ Rubrics\end{tabular}}} & \multicolumn{1}{c}{\textbf{\begin{tabular}[c]{@{}c@{}}ZCoT W/\\ Answers \&\\ LLM Rubrics\end{tabular}}} \vrule width 1pt & \multicolumn{1}{c|}{\textbf{\begin{tabular}[c]{@{}c@{}}ZCoT W/\\ Answers\end{tabular}}} & \multicolumn{1}{c|}{\textbf{\begin{tabular}[c]{@{}c@{}}ZCoT W/\\ Answers \&\\ Rubrics\end{tabular}}} & \multicolumn{1}{c}{\textbf{\begin{tabular}[c]{@{}c@{}}ZCoT W/\\ Answers \&\\ LLM Rubrics\end{tabular}}} \\ \midrule
            \multicolumn{1}{c|}{\multirow{7}{*}{{\begin{tabular}[c]{@{}c@{}}Introductory \\ Astronomy\end{tabular}}}} & \multicolumn{1}{c|}{Q1}            & \multicolumn{1}{c}{1.85} \vrule width 1pt                                                           & \multicolumn{1}{c|}{1.20}                                        & \multicolumn{1}{c|}{1.40}                                                    & \multicolumn{1}{c}{1.60} \vrule width 1pt                                                                   & \multicolumn{1}{c|}{\textbf{1.05}}                                     & \multicolumn{1}{c|}{1.30}                                                    & \multicolumn{1}{c}{1.30}                                                 \\ \cmidrule(l){2-9} 
            \multicolumn{1}{c|}{}                         & \multicolumn{1}{c|}{Q2}                                                                        & \multicolumn{1}{c}{1.25} \vrule width 1pt                                                           & \multicolumn{1}{c|}{0.80}                                        & \multicolumn{1}{c|}{0.90}                                                    & \multicolumn{1}{c}{1.00} \vrule width 1pt                                                                   & \multicolumn{1}{c|}{\textbf{0.45}}                                     & \multicolumn{1}{c|}{0.50}                                                    & \multicolumn{1}{c}{0.50}                                                 \\ \cmidrule(l){2-9} 
            \multicolumn{1}{c|}{}                         & \multicolumn{1}{c|}{Q3}                                                                        & \multicolumn{1}{c}{0.70} \vrule width 1pt                                                           & \multicolumn{1}{c|}{1.10}                                        & \multicolumn{1}{c|}{0.90}                                                    & \multicolumn{1}{c}{1.30} \vrule width 1pt                                                                   & \multicolumn{1}{c|}{0.50}                                     & \multicolumn{1}{c|}{\textbf{0.40}}                                                    & \multicolumn{1}{c}{\textbf{0.40}}                                                 \\ \cmidrule(l){2-9} 
            \multicolumn{1}{c|}{}                         & \multicolumn{1}{c|}{Q4}                                                                        & \multicolumn{1}{c}{1.15} \vrule width 1pt                                                           & \multicolumn{1}{c|}{1.00}                                        & \multicolumn{1}{c|}{0.70}                                                    & \multicolumn{1}{c}{0.80} \vrule width 1pt                                                                   & \multicolumn{1}{c|}{\textbf{0.50}}                                     & \multicolumn{1}{c|}{0.70}                                                    & \multicolumn{1}{c}{0.55}                                                 \\ \cmidrule(l){2-9} 
            \multicolumn{1}{c|}{}                         & \multicolumn{1}{c|}{Q5}                                                                        & \multicolumn{1}{c}{2.10} \vrule width 1pt                                                           & \multicolumn{1}{c|}{1.30}                                        & \multicolumn{1}{c|}{2.30}                                                    & \multicolumn{1}{c}{\textbf{1.10}} \vrule width 1pt                                                                   & \multicolumn{1}{c|}{1.30}                                     & \multicolumn{1}{c|}{1.40}                                                    & \multicolumn{1}{c}{1.25}                                                 \\ \midrule[0.9pt]
            \multicolumn{1}{c|}{\multirow{4}{*}{{\begin{tabular}[c]{@{}c@{}}Astrobiology\end{tabular}}}}   & \multicolumn{1}{c|}{Q1}                       & \multicolumn{1}{c}{1.30} \vrule width 1pt                                                           & \multicolumn{1}{c|}{1.60}                                        & \multicolumn{1}{c|}{1.30}                                                    & \multicolumn{1}{c}{1.68} \vrule width 1pt                                                                   & \multicolumn{1}{c|}{1.70}                                     & \multicolumn{1}{c|}{\textbf{0.70}}                                                    & \multicolumn{1}{c}{1.30}                                                 \\ \cmidrule(l){2-9} 
            \multicolumn{1}{c|}{}                         & \multicolumn{1}{c|}{Q2}                                                                        & \multicolumn{1}{c}{1.35} \vrule width 1pt                                                           & \multicolumn{1}{c|}{1.90}                                        & \multicolumn{1}{c|}{\textbf{1.00}}                                                    & \multicolumn{1}{c}{2.38} \vrule width 1pt                                                                   & \multicolumn{1}{c|}{1.40}                                     & \multicolumn{1}{c|}{1.10}                                                    & \multicolumn{1}{c}{1.20}                                                 \\ \cmidrule(l){2-9} 
            \multicolumn{1}{c|}{}                         & \multicolumn{1}{c|}{Q3}                                                                        & \multicolumn{1}{c}{1.85} \vrule width 1pt                                                           & \multicolumn{1}{c|}{2.20}                                        & \multicolumn{1}{c|}{2.00}                                                    & \multicolumn{1}{c}{3.02} \vrule width 1pt                                                                   & \multicolumn{1}{c|}{1.60}                                     & \multicolumn{1}{c|}{1.80}                                                    & \multicolumn{1}{c}{2.00}                                                 \\ \midrule[0.9pt]
\multicolumn{1}{c|}{\multirow{5.5}{*}{{\begin{tabular}[c]{@{}c@{}}History \& \\ Philosophy of \\ Astronomy \end{tabular}}}}     & \multicolumn{1}{c|}{Q1}  & \multicolumn{1}{c}{0.70} \vrule width 1pt                                                           & \multicolumn{1}{c|}{0.80}                                        & \multicolumn{1}{c|}{1.50}                                                    & \multicolumn{1}{c}{0.95} \vrule width 1pt                                                                   & \multicolumn{1}{c|}{\textbf{0.40}}                                     & \multicolumn{1}{c|}{0.50}                                                    & \multicolumn{1}{c}{\textbf{0.40}}                                                 \\ \cmidrule(l){2-9} 
            \multicolumn{1}{c|}{}                         & \multicolumn{1}{c|}{Q2}                                                                        & \multicolumn{1}{c}{1.30} \vrule width 1pt                                                           & \multicolumn{1}{c|}{0.90}                                        & \multicolumn{1}{c|}{1.20}                                                    & \multicolumn{1}{c}{1.04} \vrule width 1pt                                                                   & \multicolumn{1}{c|}{1.05}                                     & \multicolumn{1}{c|}{1.10}                                                    & \multicolumn{1}{c}{\textbf{0.75}}                                                 \\ \cmidrule(l){2-9} 
            \multicolumn{1}{c|}{}                         & \multicolumn{1}{c|}{Q3}                                                                        & \multicolumn{1}{c}{1.30} \vrule width 1pt                                                           & \multicolumn{1}{c|}{0.80}                                        & \multicolumn{1}{c|}{1.50}                                                    & \multicolumn{1}{c}{1.00} \vrule width 1pt                                                                   & \multicolumn{1}{c|}{0.95}                                     & \multicolumn{1}{c|}{\textbf{0.70}}                                                    & \multicolumn{1}{c}{\textbf{0.70}}                                                 \\ \cmidrule(l){2-9} 
            \multicolumn{1}{c|}{}                         & \multicolumn{1}{c|}{Q4}                                                                        & \multicolumn{1}{c}{1.40} \vrule width 1pt                                                           & \multicolumn{1}{c|}{\textbf{0.60}}                                        & \multicolumn{1}{c|}{1.30}                                                    & \multicolumn{1}{c}{\textbf{0.60}} \vrule width 1pt                                                                   & \multicolumn{1}{c|}{1.25}                                     & \multicolumn{1}{c|}{0.90}                                                    & \multicolumn{1}{c}{0.95}                                                 \\ \midrule[0.9pt]
\multicolumn{1}{c|}{Overall Average} & \multicolumn{1}{c|}{All} & \multicolumn{1}{c}{1.35} \vrule width 1pt & \multicolumn{1}{c|}{1.18} & \multicolumn{1}{c|}{1.33} & \multicolumn{1}{c}{1.37} \vrule width 1pt & \multicolumn{1}{c|}{1.01} & \multicolumn{1}{c|}{\textbf{0.92}} & \multicolumn{1}{c}{0.94} \\ \bottomrule

        \end{tabular}
    \end{adjustbox}
\end{table*}

\begin{table*}[!]
    \centering
    \caption{MAE values for 10 students' writing assignments using bootstrap resampled results with 10,000 iterations, comparing LLM-assigned and peer grades to instructor-assigned grades. \textbf{Bold} values indicate the lowest MAE for each question.}
    \label{tab:mae-bootstrap}
    \footnotesize

    \begin{adjustbox}{width=\textwidth,center}

        \begin{tabular}{@{}cccccccccc@{}}
            \toprule
                                                                                         &                                                                                        & \multicolumn{1}{l}{}                                                                & \multicolumn{3}{c}{\textbf{GPT-3.5}}                                                                                                                                                                                                                  & \multicolumn{3}{c}{\textbf{GPT-4}}                                                                                                                                                                                               \\ \cmidrule(lr){4-6}  \cmidrule(l){7-9}
            \multicolumn{1}{c|}{\textbf{Courses}}         & \multicolumn{1}{c|}{\textbf{\raisebox{-2.5ex}{\rotatebox{90}{Questions}}}}  & \multicolumn{1}{c}{\textbf{\begin{tabular}[c]{@{}c@{}}Peer\\ Grades\end{tabular}}} \vrule width 1pt & \multicolumn{1}{c|}{\textbf{\begin{tabular}[c]{@{}c@{}}ZCoT W/\\ Answers\end{tabular}}} & \multicolumn{1}{c|}{\textbf{\begin{tabular}[c]{@{}c@{}}ZCoT W/\\ Answers \&\\ Rubrics\end{tabular}}} & \multicolumn{1}{c}{\textbf{\begin{tabular}[c]{@{}c@{}}ZCoT W/\\ Answers \&\\ LLM Rubrics\end{tabular}}} \vrule width 1pt & \multicolumn{1}{c|}{\textbf{\begin{tabular}[c]{@{}c@{}}ZCoT W/\\ Answers\end{tabular}}} & \multicolumn{1}{c|}{\textbf{\begin{tabular}[c]{@{}c@{}}ZCoT W/\\ Answers \&\\ Rubrics\end{tabular}}} & \multicolumn{1}{c}{\textbf{\begin{tabular}[c]{@{}c@{}}ZCoT W/\\ Answers \&\\ LLM Rubrics\end{tabular}}} \\ \midrule
            \multicolumn{1}{c|}{\multirow{7}{*}{{\begin{tabular}[c]{@{}c@{}}Introductory \\ Astronomy\end{tabular}}}} & \multicolumn{1}{c|}{Q1}            & \multicolumn{1}{c}{1.67} \vrule width 1pt                                                           & \multicolumn{1}{c|}{\textbf{1.30}}                                        & \multicolumn{1}{c|}{2.32}                                                    & \multicolumn{1}{c}{2.06} \vrule width 1pt                                                                   & \multicolumn{1}{c|}{1.72}                                     & \multicolumn{1}{c|}{1.68}                                                    & \multicolumn{1}{c}{1.77}                                                 \\ \cmidrule(l){2-9} 
            \multicolumn{1}{c|}{}                         & \multicolumn{1}{c|}{Q2}                                                                        & \multicolumn{1}{c}{1.68} \vrule width 1pt                                                           & \multicolumn{1}{c|}{1.32}                                        & \multicolumn{1}{c|}{1.01}                                                    & \multicolumn{1}{c}{0.96} \vrule width 1pt                                                                   & \multicolumn{1}{c|}{\textbf{0.91}}                                     & \multicolumn{1}{c|}{1.02}                                                    & \multicolumn{1}{c}{0.98}                                                 \\ \cmidrule(l){2-9} 
            \multicolumn{1}{c|}{}                         & \multicolumn{1}{c|}{Q3}                                                                        & \multicolumn{1}{c}{2.43} \vrule width 1pt                                                           & \multicolumn{1}{c|}{2.69}                                        & \multicolumn{1}{c|}{2.58}                                                    & \multicolumn{1}{c}{2.85} \vrule width 1pt                                                                   & \multicolumn{1}{c|}{2.37}                                     & \multicolumn{1}{c|}{2.51}                                                    & \multicolumn{1}{c}{\textbf{2.35}}                                                 \\ \cmidrule(l){2-9} 
            \multicolumn{1}{c|}{}                         & \multicolumn{1}{c|}{Q4}                                                                        & \multicolumn{1}{c}{\textbf{2.14}} \vrule width 1pt                                                           & \multicolumn{1}{c|}{2.46}                                        & \multicolumn{1}{c|}{2.62}                                                    & \multicolumn{1}{c}{2.74} \vrule width 1pt                                                                   & \multicolumn{1}{c|}{2.34}                                     & \multicolumn{1}{c|}{2.68}                                                    & \multicolumn{1}{c}{2.62}                                                 \\ \cmidrule(l){2-9} 
            \multicolumn{1}{c|}{}                         & \multicolumn{1}{c|}{Q5}                                                                        & \multicolumn{1}{c}{3.30} \vrule width 1pt                                                           & \multicolumn{1}{c|}{\textbf{2.96}}                                        & \multicolumn{1}{c|}{3.55}                                                    & \multicolumn{1}{c}{3.59} \vrule width 1pt                                                                   & \multicolumn{1}{c|}{3.22}                                     & \multicolumn{1}{c|}{3.53}                                                    & \multicolumn{1}{c}{3.61}                                                 \\ \midrule[0.9pt]
            \multicolumn{1}{c|}{\multirow{4}{*}{{\begin{tabular}[c]{@{}c@{}}Astrobiology\end{tabular}}}}   & \multicolumn{1}{c|}{Q1}                       & \multicolumn{1}{c}{3.35} \vrule width 1pt                                                           & \multicolumn{1}{c|}{3.39}                                        & \multicolumn{1}{c|}{3.92}                                                    & \multicolumn{1}{c}{3.62} \vrule width 1pt                                                                   & \multicolumn{1}{c|}{\textbf{3.30}}                                     & \multicolumn{1}{c|}{3.57}                                                    & \multicolumn{1}{c}{3.47}                                                 \\ \cmidrule(l){2-9} 
            \multicolumn{1}{c|}{}                         & \multicolumn{1}{c|}{Q2}                                                                        & \multicolumn{1}{c}{3.40} \vrule width 1pt                                                           & \multicolumn{1}{c|}{\textbf{2.93}}                                        & \multicolumn{1}{c|}{3.03}                                                    & \multicolumn{1}{c}{4.08} \vrule width 1pt                                                                   & \multicolumn{1}{c|}{\textbf{2.93}}                                     & \multicolumn{1}{c|}{3.07}                                                    & \multicolumn{1}{c}{3.27}                                                 \\ \cmidrule(l){2-9} 
            \multicolumn{1}{c|}{}                         & \multicolumn{1}{c|}{Q3}                                                                        & \multicolumn{1}{c}{2.32} \vrule width 1pt                                                           & \multicolumn{1}{c|}{3.06}                                        & \multicolumn{1}{c|}{3.27}                                                    & \multicolumn{1}{c}{4.57} \vrule width 1pt                                                                   & \multicolumn{1}{c|}{\textbf{2.25}}                                     & \multicolumn{1}{c|}{2.51}                                                    & \multicolumn{1}{c}{2.49}                                                 \\ \midrule[0.9pt]
\multicolumn{1}{c|}{\multirow{5.5}{*}{{\begin{tabular}[c]{@{}c@{}}History \& \\ Philosophy of \\ Astronomy \end{tabular}}}}     & \multicolumn{1}{c|}{Q1}  & \multicolumn{1}{c}{\textbf{0.64}} \vrule width 1pt                                                           & \multicolumn{1}{c|}{0.94}                                        & \multicolumn{1}{c|}{1.50}                                                    & \multicolumn{1}{c}{0.87} \vrule width 1pt                                                                   & \multicolumn{1}{c|}{0.66}                                     & \multicolumn{1}{c|}{0.86}                                                    & \multicolumn{1}{c}{0.78}                                                 \\ \cmidrule(l){2-9} 
            \multicolumn{1}{c|}{}                         & \multicolumn{1}{c|}{Q2}                                                                        & \multicolumn{1}{c}{1.44} \vrule width 1pt                                                           & \multicolumn{1}{c|}{0.97}                                        & \multicolumn{1}{c|}{1.14}                                                    & \multicolumn{1}{c}{\textbf{0.94}} \vrule width 1pt                                                                   & \multicolumn{1}{c|}{1.16}                                     & \multicolumn{1}{c|}{0.96}                                                    & \multicolumn{1}{c}{1.00}                                                 \\ \cmidrule(l){2-9} 
            \multicolumn{1}{c|}{}                         & \multicolumn{1}{c|}{Q3}                                                                        & \multicolumn{1}{c}{1.28} \vrule width 1pt                                                           & \multicolumn{1}{c|}{0.82}                                        & \multicolumn{1}{c|}{1.74}                                                    & \multicolumn{1}{c}{1.32} \vrule width 1pt                                                                   & \multicolumn{1}{c|}{1.02}                                     & \multicolumn{1}{c|}{\textbf{0.75}}                                                    & \multicolumn{1}{c}{0.77}                                                 \\ \cmidrule(l){2-9} 
            \multicolumn{1}{c|}{}                         & \multicolumn{1}{c|}{Q4}                                                                        & \multicolumn{1}{c}{1.48} \vrule width 1pt                                                           & \multicolumn{1}{c|}{0.68}                                        & \multicolumn{1}{c|}{1.42}                                                    & \multicolumn{1}{c}{\textbf{0.56}} \vrule width 1pt                                                                   & \multicolumn{1}{c|}{1.21}                                     & \multicolumn{1}{c|}{1.00}                                                    & \multicolumn{1}{c}{0.99}                                                 \\ \midrule[0.9pt]
            \multicolumn{1}{c|}{Overall Average}                 & \multicolumn{1}{c|}{All}                                                                          & \multicolumn{1}{c}{2.09} \vrule width 1pt                                                           & \multicolumn{1}{c|}{1.96}                                        & \multicolumn{1}{c|}{2.34}                                                    & \multicolumn{1}{c}{2.35} \vrule width 1pt                                                                   & \multicolumn{1}{c|}{\textbf{1.92}}                                     & \multicolumn{1}{c|}{2.01}                                                    & \multicolumn{1}{c}{2.01}                                                 \\ \bottomrule
        \end{tabular}
    \end{adjustbox}
\end{table*}

\begin{table*}[!th]
    \centering
    \caption{$p$-values for results in Table \ref{tab:raw-resullts}, computed through bootstrap resampling with 10,000 iterations. None of the $p$-values exceed the 0.05 threshold for significance, implying no statistically significant difference in average grades is detected.}
    \label{tab:p_values}
    \footnotesize

    \begin{adjustbox}{width=\textwidth,center}

        \begin{tabular}{@{}cccccccccc@{}}
            \toprule
                                                                                         &                                                                                        & \multicolumn{1}{l}{}                                                                & \multicolumn{3}{c}{\textbf{GPT-3.5}}                                                                                                                                                                                                                  & \multicolumn{3}{c}{\textbf{GPT-4}}                                                                                                                                                                                               \\ \cmidrule(lr){4-6}  \cmidrule(l){7-9}
            \multicolumn{1}{c|}{\textbf{Courses}}         & \multicolumn{1}{c|}{\textbf{\raisebox{-2.5ex}{\rotatebox{90}{Questions}}}}  & \multicolumn{1}{c}{\textbf{\begin{tabular}[c]{@{}c@{}}Peer\\ Grades\end{tabular}}} \vrule width 1pt & \multicolumn{1}{c|}{\textbf{\begin{tabular}[c]{@{}c@{}}ZCoT W/\\ Answers\end{tabular}}} & \multicolumn{1}{c|}{\textbf{\begin{tabular}[c]{@{}c@{}}ZCoT W/\\ Answers \&\\ Rubrics\end{tabular}}} & \multicolumn{1}{c}{\textbf{\begin{tabular}[c]{@{}c@{}}ZCoT W/\\ Answers \&\\ LLM Rubrics\end{tabular}}} \vrule width 1pt & \multicolumn{1}{c|}{\textbf{\begin{tabular}[c]{@{}c@{}}ZCoT W/\\ Answers\end{tabular}}} & \multicolumn{1}{c|}{\textbf{\begin{tabular}[c]{@{}c@{}}ZCoT W/\\ Answers \&\\ Rubrics\end{tabular}}} & \multicolumn{1}{c}{\textbf{\begin{tabular}[c]{@{}c@{}}ZCoT W/\\ Answers \&\\ LLM Rubrics\end{tabular}}} \\ \midrule
            \multicolumn{1}{c|}{\multirow{7}{*}{{\begin{tabular}[c]{@{}c@{}}Introductory \\ Astronomy\end{tabular}}}} & \multicolumn{1}{c|}{Q1}            & \multicolumn{1}{c}{0.48} \vrule width 1pt                                                           & \multicolumn{1}{c|}{0.49}                                        & \multicolumn{1}{c|}{0.54}                                                    & \multicolumn{1}{c}{0.86} \vrule width 1pt                                                                   & \multicolumn{1}{c|}{0.49}                                     & \multicolumn{1}{c|}{0.52}                                                    & \multicolumn{1}{c}{0.55}                                                 \\ \cmidrule(l){2-9} 
            \multicolumn{1}{c|}{}                         & \multicolumn{1}{c|}{Q2}                                                                        & \multicolumn{1}{c}{0.52} \vrule width 1pt                                                           & \multicolumn{1}{c|}{0.57}                                        & \multicolumn{1}{c|}{0.78}                                                    & \multicolumn{1}{c}{0.64} \vrule width 1pt                                                                   & \multicolumn{1}{c|}{0.49}                                     & \multicolumn{1}{c|}{0.78}                                                    & \multicolumn{1}{c}{0.60}                                                 \\ \cmidrule(l){2-9} 
            \multicolumn{1}{c|}{}                         & \multicolumn{1}{c|}{Q3}                                                                        & \multicolumn{1}{c}{0.95} \vrule width 1pt                                                           & \multicolumn{1}{c|}{0.63}                                        & \multicolumn{1}{c|}{0.84}                                                    & \multicolumn{1}{c}{0.59} \vrule width 1pt                                                                   & \multicolumn{1}{c|}{0.97}                                     & \multicolumn{1}{c|}{0.89}                                                    & \multicolumn{1}{c}{0.95}                                                 \\ \cmidrule(l){2-9} 
            \multicolumn{1}{c|}{}                         & \multicolumn{1}{c|}{Q4}                                                                        & \multicolumn{1}{c}{0.98} \vrule width 1pt                                                           & \multicolumn{1}{c|}{0.64}                                        & \multicolumn{1}{c|}{0.80}                                                    & \multicolumn{1}{c}{0.67} \vrule width 1pt                                                                   & \multicolumn{1}{c|}{0.96}                                     & \multicolumn{1}{c|}{0.74}                                                    & \multicolumn{1}{c}{0.79}                                                 \\ \cmidrule(l){2-9} 
            \multicolumn{1}{c|}{}                         & \multicolumn{1}{c|}{Q5}                                                                        & \multicolumn{1}{c}{0.50} \vrule width 1pt                                                           & \multicolumn{1}{c|}{0.57}                                        & \multicolumn{1}{c|}{0.52}                                                    & \multicolumn{1}{c}{0.61} \vrule width 1pt                                                                   & \multicolumn{1}{c|}{0.66}                                     & \multicolumn{1}{c|}{0.79}                                                    & \multicolumn{1}{c}{0.64}                                                 \\ \midrule[0.9pt]
            \multicolumn{1}{c|}{\multirow{4}{*}{{\begin{tabular}[c]{@{}c@{}}Astrobiology\end{tabular}}}}   & \multicolumn{1}{c|}{Q1}                       & \multicolumn{1}{c}{0.66} \vrule width 1pt                                                           & \multicolumn{1}{c|}{0.69}                                        & \multicolumn{1}{c|}{0.98}                                                    & \multicolumn{1}{c}{0.89} \vrule width 1pt                                                                   & \multicolumn{1}{c|}{0.65}                                     & \multicolumn{1}{c|}{0.87}                                                    & \multicolumn{1}{c}{0.86}                                                 \\ \cmidrule(l){2-9} 
            \multicolumn{1}{c|}{}                         & \multicolumn{1}{c|}{Q2}                                                                        & \multicolumn{1}{c}{0.65} \vrule width 1pt                                                           & \multicolumn{1}{c|}{0.84}                                        & \multicolumn{1}{c|}{0.69}                                                    & \multicolumn{1}{c}{0.55} \vrule width 1pt                                                                   & \multicolumn{1}{c|}{0.53}                                     & \multicolumn{1}{c|}{0.64}                                                    & \multicolumn{1}{c}{0.80}                                                 \\ \cmidrule(l){2-9} 
            \multicolumn{1}{c|}{}                         & \multicolumn{1}{c|}{Q3}                                                                        & \multicolumn{1}{c}{0.50} \vrule width 1pt                                                           & \multicolumn{1}{c|}{0.51}                                        & \multicolumn{1}{c|}{0.52}                                                    & \multicolumn{1}{c}{0.50} \vrule width 1pt                                                                   & \multicolumn{1}{c|}{0.87}                                     & \multicolumn{1}{c|}{0.71}                                                    & \multicolumn{1}{c}{0.72}                                                 \\ \midrule[0.9pt]
\multicolumn{1}{c|}{\multirow{5.5}{*}{{\begin{tabular}[c]{@{}c@{}}History \& \\ Philosophy of \\ Astronomy \end{tabular}}}}     & \multicolumn{1}{c|}{Q1}  & \multicolumn{1}{c}{0.83} \vrule width 1pt                                                           & \multicolumn{1}{c|}{0.59}                                        & \multicolumn{1}{c|}{0.61}                                                    & \multicolumn{1}{c}{0.52} \vrule width 1pt                                                                   & \multicolumn{1}{c|}{1.00}                                     & \multicolumn{1}{c|}{0.61}                                                    & \multicolumn{1}{c}{0.57}                                                 \\ \cmidrule(l){2-9} 
            \multicolumn{1}{c|}{}                         & \multicolumn{1}{c|}{Q2}                                                                        & \multicolumn{1}{c}{0.50} \vrule width 1pt                                                           & \multicolumn{1}{c|}{0.56}                                        & \multicolumn{1}{c|}{0.55}                                                    & \multicolumn{1}{c}{0.71} \vrule width 1pt                                                                   & \multicolumn{1}{c|}{0.52}                                     & \multicolumn{1}{c|}{0.52}                                                    & \multicolumn{1}{c}{0.51}                                                 \\ \cmidrule(l){2-9} 
            \multicolumn{1}{c|}{}                         & \multicolumn{1}{c|}{Q3}                                                                        & \multicolumn{1}{c}{0.58} \vrule width 1pt                                                           & \multicolumn{1}{c|}{1.00}                                        & \multicolumn{1}{c|}{0.49}                                                    & \multicolumn{1}{c}{0.48} \vrule width 1pt                                                                   & \multicolumn{1}{c|}{0.55}                                     & \multicolumn{1}{c|}{0.57}                                                    & \multicolumn{1}{c}{0.55}                                                 \\ \cmidrule(l){2-9} 
            \multicolumn{1}{c|}{}                         & \multicolumn{1}{c|}{Q4}                                                                        & \multicolumn{1}{c}{0.56} \vrule width 1pt                                                           & \multicolumn{1}{c|}{0.54}                                        & \multicolumn{1}{c|}{0.53}                                                    & \multicolumn{1}{c}{1.00} \vrule width 1pt                                                                   & \multicolumn{1}{c|}{0.51}                                     & \multicolumn{1}{c|}{0.55}                                                    & \multicolumn{1}{c}{0.51}                                                 \\ \bottomrule
        \end{tabular}
    \end{adjustbox}
\end{table*}

\section{Details on $p$-values}
\label{appendix:p_values}

As detailed in Section \ref{section:experimental-setup} under Bootstrap Resampling, we use 10,000 iterations with replacement in all settings with bootstrap resampling. Table \ref{tab:p_values} presents the $p$-values for the differences in average scores referenced in Table \ref{tab:raw-resullts}. Note that the significance of average differences is assessed using a threshold of 0.05. Based on the computed $p$-values, none of the differences in average scores is statistically significant.

\end{document}